\documentclass[pdflatex,sn-mathphys-num]{sn-jnl}

\usepackage{xcolor}

\usepackage{amsmath,amssymb}

\usepackage{graphicx}

\usepackage[section]{placeins}

\usepackage{booktabs}
\usepackage{array}

\usepackage{siunitx}

\usepackage{tikz}
\usetikzlibrary{arrows.meta,positioning,fit,shapes.geometric,calc}
\usepackage{pgfplots}
\pgfplotsset{compat=1.18}

\usepackage{xspace}

\usepackage[capitalise,noabbrev]{cleveref}

\usepackage{pifont}

\let\orcidlogo\relax
\usepackage{orcidlink}
\DeclareSIUnit{\px}{px}

\newcommand{\cmark}{\ding{51}}
\newcommand{\xmark}{\ding{55}}

\begin{document}

\title[Planner--Controller Interaction for Ackermann Navigation]
{Planner--Controller Interaction for Ackermann Navigation: A Matched Simulation and Hardware Benchmark}

\author*[1,2]{\fnm{Moaiz} \sur{Saeed}\,\orcidlink{0009-0002-1005-286X}}\email{ms2277@hw.ac.uk}

\author[1]{\fnm{Marah} \sur{Saleh}\,\orcidlink{0009-0008-4832-2750}}

\author[2]{\fnm{Mounis} \sur{Shawgi}\,\orcidlink{0000-0003-2350-7640}}

\author[1]{\fnm{Mohamed} \sur{Al Musleh}\,\orcidlink{0009-0001-5774-564X}}

\affil*[1]{\orgdiv{AIMBeyonD Lab},
\orgname{Heriot-Watt University Dubai},
\orgaddress{\city{Dubai}, \country{United Arab Emirates}}}

\affil[2]{\orgdiv{School of Engineering and Physical Sciences},
\orgname{Heriot-Watt University Dubai},
\orgaddress{\city{Dubai}, \country{United Arab Emirates}}}

\abstract{
Global planners and path-tracking controllers are often evaluated independently, although their interaction determines how a vehicle follows a planned path. Whether simulation accurately predicts the performance of specific planner--controller combinations on Ackermann-steered hardware remains uncertain. This study evaluated three geometric controllers (Regulated Pure Pursuit, Stanley, and Vector Pursuit) paired with two global planners (Hybrid A* and NavFn) on a 1/10-scale Quanser QCar and in a matched Gazebo simulation. Both environments used shared custom planning and control algorithms, matched route geometry and goals, ground-truth pose references, and a common evaluation pipeline. Five runs per configuration and environment yielded 60 benchmark runs. Controller rankings for tracking accuracy and command smoothness transferred consistently between environments, with Spearman correlations of 0.77--0.94. However, planner-dependent effects showed substantially weaker agreement. In simulation, planner selection changed the root-mean-square (RMS) angular acceleration of the commanded velocities by no more than \SI{0.030}{\radian\per\second\squared}, whereas Hybrid A* increased it by \SIrange{0.141}{0.843}{\radian\per\second\squared} on hardware, depending on the controller. These differences coincided with planner-dependent achieved speeds that the simulation's idealised longitudinal response did not reproduce. Mean controller computation times remained below \SI{0.5}{\milli\second} on the Jetson TX2. The findings demonstrate that agreement in controller rankings does not guarantee accurate prediction of planner-dependent performance for specific controller pairings on hardware, highlighting the importance of physical validation.
}


\pacs[Categories]{{(3), (5)}}
\maketitle

\section{Introduction}
\label{sec:intro}

Autonomous navigation systems separate global planning, which determines a path to the goal, from local control, which generates commands to track that path. The Nav2 framework of the Robot Operating System 2 (ROS~2) provides a modular architecture in which planners and controllers can be selected independently \cite{macenski2020marathon}. However, their interaction is particularly important for Ackermann-steered vehicles, which cannot turn in place and are constrained by a minimum turning radius. Grid-based planners may generate paths containing curvature that the vehicle cannot follow \cite{macenski2026smac}, whereas kinematically constrained planners such as Hybrid A* account for vehicle turning limits during planning \cite{dolgov2008hybridastar,dolgov2010semistructured}. Consequently, the global planner determines not only the route but also the tracking demands imposed on the controller.

Geometric path-tracking controllers, including pure pursuit and Stanley, remain widely used because of their low computational requirements and relatively simple formulations \cite{rokonuzzaman2021pathtracking}. However, their responses to planned path geometry differ. Lookahead-based controllers attenuate local path variations, while controllers incorporating direct cross-track and heading-error feedback respond more closely to the local path geometry. Planner selection may therefore influence each controller differently. The Nav2 maintainers have also recognised this coupling, recommending that path smoothing and controller selection be considered together \cite{macenski2023desks}. Despite this, the influence of global planner selection on geometric controller performance, particularly under physical Ackermann steering constraints, remains insufficiently characterised.

Previous evaluations have predominantly compared controllers using fixed paths \cite{patnaik2020pathtrackers,dominguez2016lateralcontrollers,kong2024purepursuitmpc} or global planners using a fixed controller \cite{borse2024robotinonavstack}. Filotheou et al. \cite{filotheou2020rosplanners} evaluated combinations of global and local planners on a differential-drive platform, but the physical environment was not matched to the simulated environment. Although small-scale vehicles provide a practical intermediate platform between simulation and full-scale deployment \cite{caleffi2024smallscale}, simulation does not necessarily preserve controller performance rankings on hardware \cite{kadian2020sim2real,kong2024purepursuitmpc}. Existing comparisons therefore leave two related questions unresolved: whether the effect of global planner selection depends on the controller and whether these planner--controller interactions transfer from simulation to hardware. Addressing both requires matched experiments that control the planning and control implementations, route geometry, localisation reference, and evaluation procedure.

This study investigates these questions through a full-factorial benchmark of three geometric controllers (Regulated Pure Pursuit, Stanley, and Vector Pursuit) paired with two global planners (Hybrid A* and NavFn). Experiments were conducted on a 1/10-scale Quanser QCar equipped with an NVIDIA Jetson TX2 \cite{delatrinidadrendon2025platforms} and in a matched Gazebo simulation. The hardware and simulation environments used ROS~2 Dashing and Jazzy, respectively, while sharing the custom planning and control algorithms, parameters, route geometry, and key performance indicator (KPI) evaluation pipeline. Both environments employed ground-truth pose references to minimise the influence of localisation uncertainty. Five runs were conducted for each planner--controller combination in both environments, yielding 60 benchmark runs. Tracking accuracy, command smoothness, mission efficiency, and embedded computational performance were evaluated.

The study addresses three research questions:

\begin{itemize}
    \item[RQ1:] How does global planner selection influence tracking accuracy and command smoothness, and to what extent does this influence depend on the controller?
    \item[RQ2:] Which controller rankings and planner--controller effects transfer from simulation to hardware across the evaluated KPIs?
    \item[RQ3:] What computational resources do the planning and control algorithms require on the embedded platform?
\end{itemize}

The principal contributions are:

\begin{enumerate}
    \item A matched simulation--hardware benchmark of six planner--controller configurations on an Ackermann-steered vehicle, using shared custom algorithms and evaluation procedures with documented differences between environments.

    \item A quantitative analysis of planner effects, controller effects, and their interactions, supported by bootstrap confidence intervals, demonstrating substantially larger planner-dependent command variations on hardware than in simulation.

    \item An evaluation of simulation--hardware performance transfer using controller rankings, Spearman correlations, and planner-effect directions, showing that consistent controller rankings do not guarantee reliable prediction of individual planner--controller interactions.

    \item An assessment of computational performance on the Jetson TX2, accompanied by publicly available navigation software, simulation resources, analysis scripts, and processed KPI results.
\end{enumerate} \section{Related Work}\label{sec:related}
\subsection{Comparisons of path-tracking controllers}
\label{sec:rw_controllers}

Geometric path trackers such as pure pursuit and Stanley are widely used on research vehicles due to their simplicity, limited parameterisation, and low computational cost \cite{rokonuzzaman2021pathtracking}. Comparative studies have predominantly evaluated controllers against fixed reference paths. Patnaik et al. \cite{patnaik2020pathtrackers} compared pure pursuit, Stanley, proportional--integral--derivative (PID), linear--quadratic regulator (LQR), and model predictive control (MPC) controllers in simulation and on an instrumented passenger car at \SI{10}{\kilo\metre\per\hour}, using straight, circular, lane-change, and sinusoidal paths. They attributed discrepancies between simulation and hardware to steering response delays, the inability to execute small steering adjustments, and unmodelled steering and tyre dynamics, but reported neither trial counts nor quantitative cross-environment comparisons. Dominguez et al. \cite{dominguez2016lateralcontrollers} evaluated pure pursuit, Stanley, a simplified kinematic controller, and a lateral-speed controller on a physical vehicle and in simulation along a longer recorded path at speeds up to \SI{90}{\kilo\metre\per\hour}. Stanley achieved slightly higher accuracy on hardware but became unstable above \SI{25}{\kilo\metre\per\hour} in simulation with its low-speed parameters, requiring separate tuning. Passenger comfort was assessed qualitatively, with Stanley's rapid steering movements perceived as unsafe. Rokonuzzaman et al. \cite{rokonuzzaman2021pathtracking} evaluated seven controllers in simulation using a common genetic-algorithm tuning objective to ensure equal tuning effort. Stanley exhibited the highest root-mean-square (RMS) and maximum lateral errors at low speeds.

Similar evaluation approaches were adopted on scaled and indoor platforms. Kong et al. \cite{kong2024purepursuitmpc} compared pure pursuit, MPC, and model predictive contouring control (MPCC) along a precomputed minimum-curvature raceline in the F1TENTH Gym simulator and on a 1/10-scale F1TENTH vehicle. Simulation rankings did not transfer to hardware, where pure pursuit performed best, which the authors attributed to computational delays in the optimisation-based controllers. Within Nav2, Elbouhy et al. \cite{elbouhy2025localplanners} compared the DWB (dynamic window), Model Predictive Path Integral (MPPI), and Regulated Pure Pursuit (RPP) controllers on a Clearpath Jackal across identical simulated and physical obstacle scenarios, using a single A* global planner. Sant'Anna et al. \cite{santanna2023ros2controllers} evaluated the same three controllers on a TurtleBot3, exclusively in physical environments. Neither platform employed Ackermann steering. Across these studies, reference paths were predefined, recorded, precomputed, or generated by a single global planner. Consequently, the influence of planner selection on individual controller performance remained unexamined.
\subsection{Global planner selection and controller coupling}
\label{sec:rw_planners}

Grid-based planners such as NavFn assume holonomic motion, potentially generating paths that Ackermann-steered vehicles cannot follow exactly \cite{macenski2026smac,macenski2023desks}. Hybrid A* \cite{dolgov2008hybridastar,dolgov2010semistructured} addresses this limitation by maintaining continuous vehicle states within a discretised $(x,y,\theta)$ search space, producing kinematically feasible paths. Accordingly, the ROS~2 navigation maintainers recommend Hybrid A* for Ackermann vehicles and advise selecting path smoothers in conjunction with path-tracking controllers \cite{macenski2023desks}. However, these recommendations did not quantify the influence of planner selection on controller performance.

Experimental comparisons of global planners have generally held the controller fixed or evaluated planner--controller combinations without isolating their individual effects. Borse et al. \cite{borse2024robotinonavstack} compared NavFn, Smac 2D, and Theta* with a fixed MPPI controller on an omnidirectional Robotino, conducting five runs per path in Webots and on hardware. Hardware experiments used GPS-derived poses instead of simulated odometry, comparable to the ground-truth reference employed here (Section~\ref{sec:track}). Hardware execution time was approximately 1.01 times that of simulation. Filotheou et al. \cite{filotheou2020rosplanners} evaluated nine combinations of three global and three local ROS 1 planners on a differential-drive TurtleBot 2, conducting ten runs per combination across two simulated environments and one physical laboratory. However, the laboratory was not replicated in simulation, and results from all three environments were aggregated into a single composite score. Consequently, the study ranked planner combinations without assessing whether their performance transferred from simulation to hardware. Neither Ackermann steering nor geometric path trackers were considered.

\subsection{Simulation-to-hardware performance agreement}
\label{sec:rw_agreement}

Small-scale vehicles bridge the gap between simulation, which often lacks physical fidelity, and full-scale testing, which is constrained by time, space, and cost \cite{caleffi2024smallscale}. However, a systematic review of 38 studies found that simulation-to-hardware discrepancies were rarely examined in depth, with transferability generally reported without detailed quantitative analysis \cite{caleffi2024smallscale}. Where agreement was assessed, it was often limited to individual metrics, such as execution-time ratios \cite{borse2024robotinonavstack}, or qualitative observations of vehicle behaviour \cite{patnaik2020pathtrackers}.

More detailed approaches have examined signal-level agreement and performance rankings. García Daza et al. \cite{garciadaza2023simtoreal} replicated a physical route in simulation and quantified the reality gap of an MPC path-tracking controller using Pearson correlation and maximum normalised cross-correlation of lateral error, steering angle, lateral acceleration, and lateral jerk. While this assessed behavioural agreement for an individual controller, it did not establish whether relative performance rankings transferred. Kadian et al. \cite{kadian2020sim2real} addressed this distinction by defining a simulation-to-hardware correlation coefficient based on Pearson correlation across multiple methods. Evaluating nine navigation models on a LoCoBot and a scanned simulation replica of the laboratory, they reported a success-rate correlation of 0.18 and nine pairwise ranking reversals. After simulator parameter tuning, the correlation increased to 0.844. Kong et al. \cite{kong2024purepursuitmpc} similarly observed controller ranking reversals on a scaled vehicle (Section~\ref{sec:rw_controllers}).

The present study adopts a ranking-based evaluation because absolute KPI values were not directly comparable between environments (Section~\ref{sec:sim}). Agreement is assessed per KPI through rank correlation across six planner--controller combinations and further examined through the direction of each controller's planner effect (Section~\ref{sec:stats}).

\subsection{Evaluation of command smoothness}
\label{sec:rw_smoothness}

Previous studies have assessed navigation smoothness through trajectory geometry \cite{filotheou2020rosplanners,elbouhy2025localplanners}, actuation magnitude \cite{santanna2023ros2controllers}, or qualitative passenger feedback \cite{dominguez2016lateralcontrollers}. Urrea and Valencia-Aragon \cite{urrea2026localcontrollers} instead employed command-based metrics, including the integrals of squared linear and angular velocity commands and the number of angular velocity sign changes. However, their comparison of DWB, RPP, and MPPI was limited to simulation on a differential-drive platform with NavFn as the sole global planner. The reviewed literature therefore does not establish how planner selection influences the rate of change of velocity commands across geometric Ackermann controllers, or whether their smoothness rankings transfer from simulation to hardware.

\subsection{Research gap}
\label{sec:gap}

Previous studies have predominantly compared controllers along fixed paths, evaluated planners with a fixed controller, or ranked planner--controller combinations without isolating their effects. Simulation--hardware agreement has been assessed through qualitative observations, individual performance metrics, signal-level correlations, or controller rankings. To the best of our knowledge, no study has systematically evaluated multiple geometric controllers and global planners on an Ackermann-steered vehicle under matched simulation and hardware conditions, while examining the transfer of planner effects, controller rankings, and their interaction. This study addresses these gaps using matched routes, goals, ground-truth pose references, shared planning and control algorithms, and a common KPI evaluation pipeline. Command smoothness and onboard computational load are also evaluated.
 \section{System and Platform}\label{sec:platform}

\subsection{The QCar platform}
\label{sec:qcar}

\begin{figure}[!tbp]
\centering
\includegraphics[width=\linewidth]{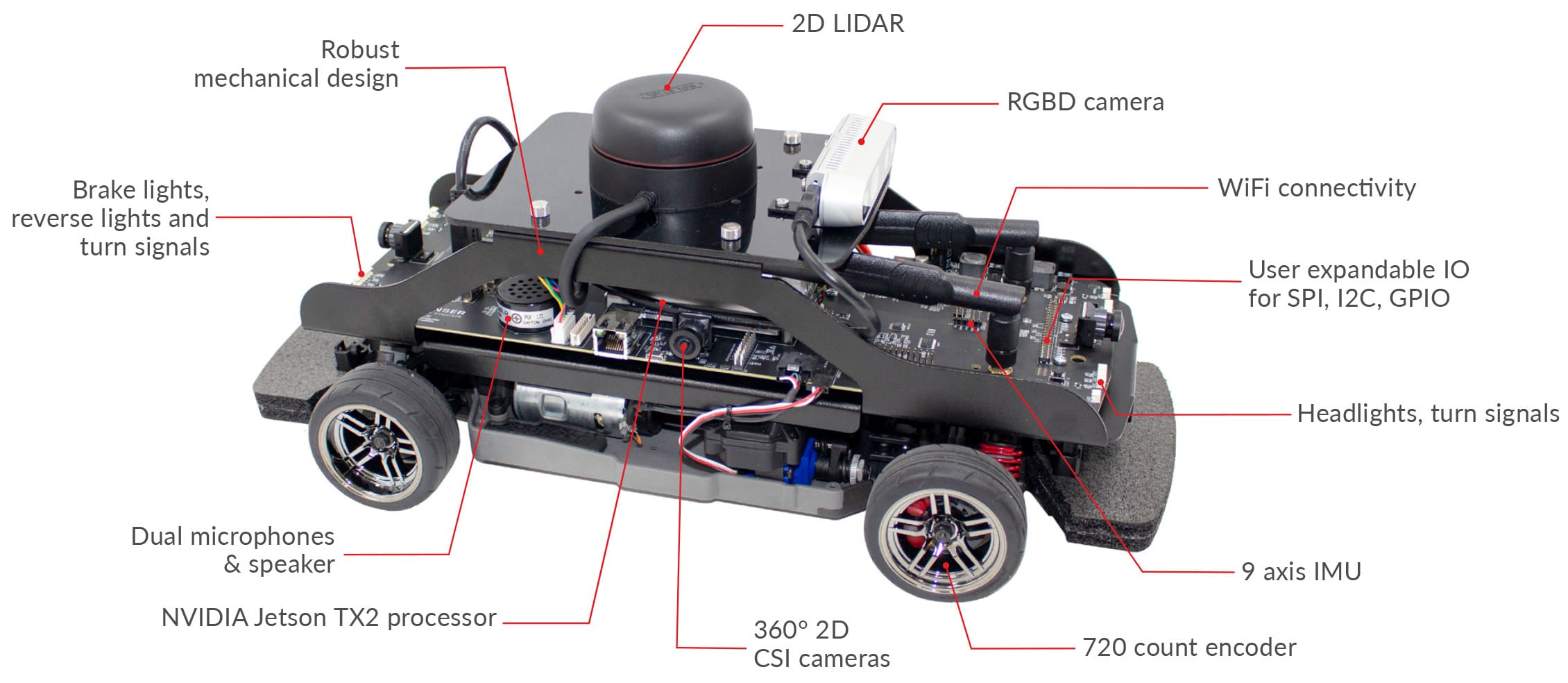}
\caption{The Quanser QCar platform with its main components}\label{fig:qcar}
\end{figure}

The Quanser QCar is a 1/10-scale Ackermann-steered research vehicle \cite{delatrinidadrendon2025platforms} (Fig.~\ref{fig:qcar}). The complete navigation stack operated on the onboard NVIDIA Jetson TX2, a resource-constrained embedded computer \cite{nvidia2017jetsontx2}.

The RPLiDAR A2 light detection and ranging (LiDAR) sensor was the only onboard sensor used, providing data for map construction and costmap obstacle layers. Vehicle pose was obtained from an external motion-capture system (Section~\ref{sec:track}).

Table~\ref{tab:qcar_params} summarises the platform parameters. The maximum steering angle of $\pm\SI{30}{\degree}$ corresponds to a minimum turning radius of approximately \SI{0.44}{\metre} ($L/\tan\delta_{\max}$), while the steering servo exhibits a first-order response with a time constant of \SI{0.16}{\second}~\cite{delatrinidadrendon2025platforms}. These constraints govern path feasibility and tracking performance.

\begin{table}[!tbp]
\caption{QCar physical parameters}
\label{tab:qcar_params}
\begin{tabular}{@{}lll@{}}
\toprule
Parameter & Symbol & Value \\
\midrule
Wheelbase              & $L$       & \SI{0.256}{\metre} \\
Wheel radius           & $r_{\mathrm{w}}$ & \SI{0.033}{\metre} \\
Encoder resolution     & $N_{\mathrm{enc}}$ & \num{28800} counts/rev (quadrature) \\
Steering time constant & $\tau$    & \SI{0.16}{\second} \\
Maximum steering angle & $\delta_{\max}$ & $\pm\SI{0.5236}{\radian}$ ($\pm\SI{30}{\degree}$) \\
\bottomrule
\end{tabular}
\end{table}

\subsection{Software stack}
\label{sec:software}

The QCar operated on Ubuntu 18.04 with ROS~2 Dashing, constrained by the Jetson TX2, whereas simulation used Ubuntu 24.04, ROS~2 Jazzy, and Gazebo Harmonic. Both environments employed Nav2 \cite{macenski2020marathon}, whose navigation pipeline is detailed in Section~\ref{sec:nav}.

The Hybrid A* planner and three controllers (RPP, Stanley, and Vector Pursuit) were developed from a common source tree compatible with both distributions. Hybrid A* operated as a standalone node implementing the \texttt{ComputePathToPose} action, while NavFn used the stock Nav2 planner server. Since Dashing predates the plugin-based controller server, each controller was hosted in a standalone action server exposing \texttt{FollowPath}, with the same servers ported to Jazzy for simulation. All controllers operated at \SI{20}{\hertz} with a position-only goal tolerance of \SI{0.15}{\metre}. Cross-environment software and configuration equivalence is detailed in Section~\ref{sec:equiv}.

\subsection{Test track, mapping, and pose reference}
\label{sec:track}

\begin{figure}[!tbp]
\centering
\includegraphics[width=\linewidth]{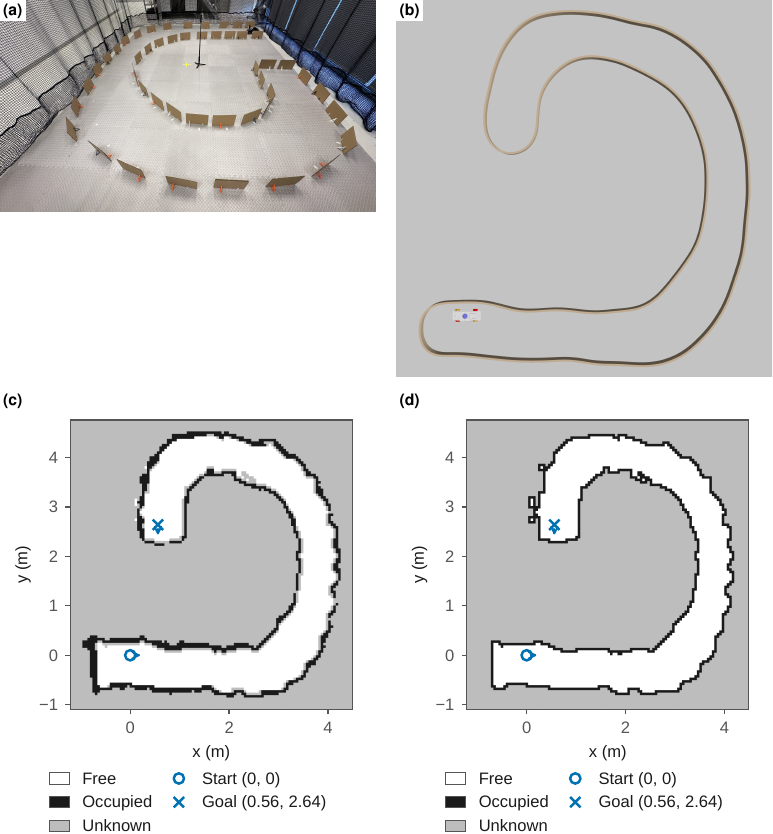}
\caption{The hook-shaped test track, bounded by cardboard walls (a), and the Gazebo track world rebuilt from it (b), with the occupancy grids used for navigation on hardware (c) and in simulation (d)}\label{fig:track}
\end{figure}

Hardware experiments were conducted on a hook-shaped corridor constructed from cardboard walls within a $\SI{5.2}{\metre} \times \SI{5.35}{\metre}$ laboratory area (Fig.~\ref{fig:track}). The median free corridor width was \SI{0.67}{\metre}. The route comprised an initial straight, a \SI{200}{\degree} left-hand bend with a centreline radius of approximately \SI{1.9}{\metre}, and a return section terminating in a tight left turn into a dead end. Although the start and goal were separated by only \SI{2.7}{\metre}, the planned route was approximately \SI{10.6}{\metre} long. The final turn, with a radius of \SI{0.5}{\metre}, approached the vehicle's minimum turning radius of \SI{0.44}{\metre}, making it the most kinematically demanding section.

Vehicle pose was obtained from an OptiTrack motion-capture system comprising six cameras tracking six rigid-body markers mounted on the QCar \cite{optitrack2026}. These measurements served as the sole odometry source, without encoder or inertial measurement unit (IMU) fusion. An occupancy grid was generated using Cartographer \cite{hess2016cartographer} from LiDAR scans and motion-capture odometry collected during teleoperation. The resulting \SI{0.05}{\metre} resolution map remained unchanged throughout the experiments.

\subsection{Navigation architecture}
\label{sec:nav}

\begin{figure}[!tbp]
\centering
\begin{tikzpicture}[
    font=\sffamily\fontsize{8}{9.6}\selectfont,
    >={Stealth[length=4pt]},
    every node/.style={inner sep=2pt},
    comp/.style={draw=black, rectangle, align=center, minimum height=9mm},
    data/.style={draw=black, fill=black!12, rectangle, align=center, minimum height=9mm},
    aux/.style={draw=black!55, rectangle, align=center, minimum height=5.5mm, text=black!75},
    lab/.style={align=center, text=black!75},
    flow/.style={->, black},
    feed/.style={->, black!55},
    node distance=2.5mm,
]
\node[data] (goal) {Goal\\pose};
\node[comp, right=of goal] (bt) {Behaviour-\\tree\\navigator};
\node[comp, right=13mm of bt] (plan) {Planner\\(Hybrid A*\\or NavFn)};
\node[data, right=of plan] (gplan) {Global\\plan};
\node[comp, right=11mm of gplan] (ctrl) {Controller\\(RPP, Stanley,\\or VP)};
\node[data, right=of ctrl] (cmd) {$(v,\omega)$\\at 20\,Hz};
\node[comp, right=of cmd] (veh) {Vehicle};

\draw[flow] (goal) -- (bt);
\draw[flow] (bt) -- node[lab, above=1pt] {Compute-\\PathTo-\\Pose} (plan);
\draw[flow] (plan) -- (gplan);
\draw[flow] (gplan) -- node[lab, above=1pt] {Follow-\\Path} (ctrl);
\draw[flow] (ctrl) -- (cmd);
\draw[flow] (cmd) -- (veh);

\node[aux, above=7mm of plan] (gcm) {Global costmap};
\node[aux, above=7mm of ctrl] (lcm) {Local costmap};
\draw[feed] (gcm) -- (plan);
\draw[black!55] (lcm) -- (ctrl);

\node[aux, below=12mm of bt] (map) {map};
\node[aux, below=12mm of plan] (odom) {odom};
\node[aux, below=12mm of ctrl] (base) {base};
\draw[feed] (map) -- node[lab, below=1pt] {constant\\transform} (odom);
\draw[feed] (odom) -- node[lab, below=1pt] {motion capture /\\simulator pose} (base);

\draw[feed] (base.north) -- node[lab, right=1pt, pos=0.45] {pose} (ctrl.south);
\draw[feed] ([xshift=-3mm]base.north) |- ([yshift=-4.5mm]plan.south) -- (plan.south);
\end{tikzpicture}
\caption{Navigation pipeline, identical in both environments: one goal, one global plan, and no replanning or recovery. VP: Vector Pursuit}\label{fig:pipeline}
\end{figure}
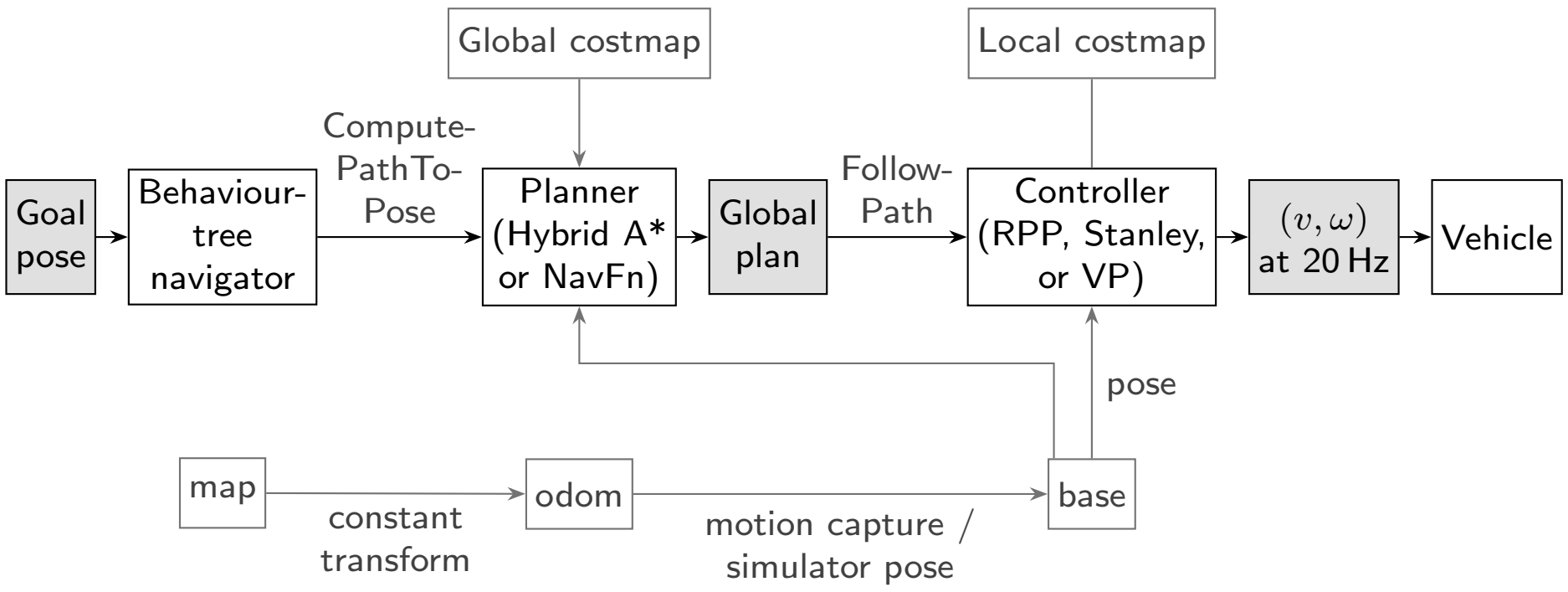

Navigation followed the standard Nav2 architecture \cite{macenski2020marathon} (Fig.~\ref{fig:pipeline}). For each run, the behaviour-tree navigator requested a global plan through the \texttt{ComputePathToPose} action and passed it to the selected controller through \texttt{FollowPath}. Replanning and recovery behaviours were disabled, ensuring that each run evaluated one controller tracking a single global plan.

Both planners exposed the same planning interface, with the generated path serving as the sole handoff to the controller. The global costmaps comprised static (Cartographer), obstacle (LiDAR), and inflation layers. Hybrid A* maintained a separate costmap instance, while NavFn used the Nav2 planner server's costmap, with identical configurations. The controllers employed rolling local costmaps with obstacle and inflation layers. All costmaps used a resolution of \SI{0.05}{\metre}, a circular vehicle footprint of radius \SI{0.24}{\metre}, and an inflation radius of \SI{0.55}{\metre}.

The transformation chain comprised map $\rightarrow$ odom $\rightarrow$ base. On hardware, the odom $\rightarrow$ base transform was obtained directly from motion capture at \SI{50}{\hertz}, without filtering or latency compensation, and timestamped upon arrival. A fixed map $\rightarrow$ odom transform aligned the Cartographer map with the motion-capture reference frame throughout the trials.

\subsection{Simulation environment}
\label{sec:sim}

\begin{figure*}[!tbp]
\centering
\includegraphics[width=\textwidth]{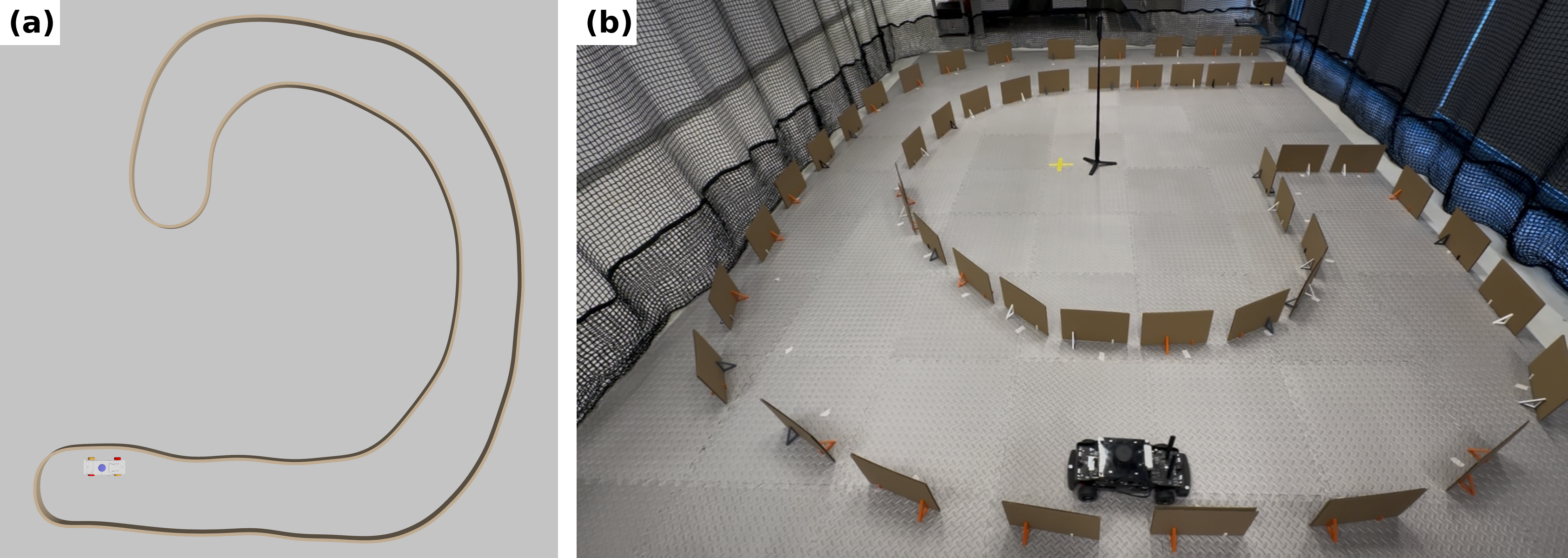}
\caption{The simulated QCar and track in Gazebo Harmonic (a) and the physical setup (b)}\label{fig:sim_hw}
\end{figure*}

The hardware configuration was replicated in Gazebo Harmonic \cite{koenig2004gazebo} (Fig.~\ref{fig:sim_hw}), using a Unified Robot Description Format (URDF) vehicle model with the physical parameters listed in Table~\ref{tab:qcar_params}. Gazebo's Ackermann steering system converted linear velocity and steering commands into wheel and steering-joint actuation, incorporating the physical steering lag ($\tau = \SI{0.16}{\second}$). However, longitudinal dynamics were idealised, with the simulated vehicle tracking commanded speeds exactly, whereas the hardware employed open-loop throttle control and exceeded commanded speeds. Consequently, absolute KPI values were not compared between environments (Section~\ref{sec:stats}).

The simulated odom $\rightarrow$ base transform was obtained from ground-truth pose at approximately \SI{48}{\hertz}, comparable to the hardware motion-capture rate. Unlike simulation, hardware measurements incurred transport latency, with mean retrieval times of \SIrange{8.3}{8.9}{\milli\second} and 95th percentiles of \SIrange{13.7}{17.4}{\milli\second}. These measurements exclude internal motion-capture processing and therefore represent lower bounds on latency. At hardware speeds of \SIrange{0.40}{0.67}{\metre\per\second}, the corresponding positional lag was approximately \SIrange{3}{12}{\milli\metre}, small relative to the corridor width and measured cross-track errors (\SIrange{34}{144}{\milli\metre}). The pose references were therefore considered sufficiently comparable for benchmarking. The simulated LiDAR matched the RPLiDAR A2 scan rate (\SI{10}{\hertz}) and range (\SIrange{0.15}{12}{\metre}), supplying only the costmap obstacle layer.

The Gazebo track geometry and navigation map were reconstructed from the hardware occupancy grid. Goal positions differed by less than \SI{1}{\centi\metre}, while the mean hardware start position was within \SI{0.08}{\metre} of the simulated start, with a standard deviation of \SI{0.03}{\metre} across runs. Hybrid A* generated plans of \SI{10.72}{\metre} in simulation and \SI{10.61}{\metre} on hardware, a difference of \SI{0.11}{\metre}. NavFn produced a larger discrepancy of \SI{0.40}{\metre}, with corresponding lengths of \SI{11.04}{\metre} and \SI{10.64}{\metre}.

\subsection{Simulation--hardware correspondence}
\label{sec:equiv}

Table~\ref{tab:equivalence} summarises the correspondence between the hardware and simulation configurations. The custom Hybrid A* planner and controller algorithms used byte-identical source files and parameters, with only their ROS~2 server wrappers adapted between Dashing and Jazzy. NavFn used the stock implementation from each distribution with matched parameters. Although the LiDAR scan resolutions differed, neither environment's obstacle layer introduced occupied cells beyond the static map. Consequently, this difference did not alter the obstacle information available to the planners and controllers.

\begin{table}[!tbp]
\caption{Simulation--hardware software and configuration correspondence}
\label{tab:equivalence}
\begin{tabular}{@{}>{\raggedright\arraybackslash}p{0.17\textwidth}
                  >{\raggedright\arraybackslash}p{0.30\textwidth}
                  >{\raggedright\arraybackslash}p{0.26\textwidth}
                  >{\raggedright\arraybackslash}p{0.15\textwidth}@{}}
\toprule
Component & Hardware & Simulation & Status \\
\midrule

OS / ROS~2
& Ubuntu 18.04 / Dashing
& Ubuntu 24.04 / Jazzy
& Differs \\

Compute
& Jetson TX2
& Desktop PC
& Differs \\

Hybrid A* planner
& Standalone node
& Same algorithm and parameters
& Identical \\

NavFn planner
& Dashing stock implementation
& Jazzy stock implementation; same parameters
& Matched parameters \\

Controllers (RPP, Stanley, Vector Pursuit)
& Standalone action servers
& Same algorithms and parameters; wrappers ported to Jazzy
& Identical algorithms \\

Navigation settings
& \SI{20}{\hertz}; single plan; no replanning or recovery; \SI{0.15}{\metre} position-only goal tolerance
& Same
& Identical \\

Global costmap
& Static, obstacle and inflation layers; \SI{0.05}{\metre} resolution; $r=\SI{0.24}{\metre}$; \SI{0.55}{\metre} inflation
& Same
& Identical \\

Pose source (odom $\rightarrow$ base)
& Motion capture, \SI{50}{\hertz}, timestamped on arrival
& Simulator ground truth, ${\approx}\,\SI{48}{\hertz}$
& Comparable \\

map $\rightarrow$ odom
& Constant offset
& Identity
& Frame-aligned \\

Navigation map
& Cartographer occupancy grid
& Generated from the same grid
& Matched \\

LiDAR
& 384 samples, ${\approx}\,\SI{10}{\hertz}$, \SIrange{0.15}{12}{\metre}
& 720 samples, \SI{10}{\hertz}, \SIrange{0.15}{12}{\metre}
& Comparable \\

Start and goal
& Hand-placed start; goal $(0.570, 2.643)$
& Start $(0, 0)$; goal $(0.56, 2.64)$
& Matched geometry \\

Steering conversion, limit and lag
& $\delta=\arctan(L\omega/v)$; $\pm\SI{0.5}{\radian}$; $\tau=\SI{0.16}{\second}$
& Same
& Identical \\

Longitudinal response
& Open-loop throttle
& Exact commanded speed
& Differs \\

\bottomrule
\end{tabular}
\end{table}  \section{Planning and Control Methods}\label{sec:methods}

\subsection{Hybrid A* and NavFn global planners}
\label{sec:planners}

NavFn and Hybrid A* were selected to compare global planning with and without explicit vehicle kinematic constraints. NavFn generates grid-based paths without enforcing curvature limits, leaving the controller to accommodate potentially infeasible turns. In contrast, Hybrid A* incorporates the vehicle's minimum turning radius during planning, reducing the burden on the controller.

NavFn constructs a navigation function through wavefront expansion from the goal and extracts a path using gradient descent \cite{konolige2000gradient}. It was configured with Dijkstra expansion, a goal tolerance of \SI{0.15}{\metre}, and planning through unknown space enabled. The planner represents the vehicle as a circle without imposing curvature or heading constraints.

Hybrid A*, based on \cite{dolgov2008hybridastar,dolgov2010semistructured}, searches over continuous states $(x,y,\theta)$ using fixed-length bicycle-model motion primitives:
\begin{equation}\label{eq:hybrid_step}
\begin{aligned}
\theta_{j+1} &= \theta_j + \frac{s}{R_{\min}}\,\sigma,\\
x_{j+1} &= x_j + s\cos\theta_{j+1},\\
y_{j+1} &= y_j + s\sin\theta_{j+1}
\end{aligned}
\end{equation}
where $j$ is the primitive step index, $\sigma \in \{-1,-0.5,0,0.5,1\}$ represents steering input as a fraction of full lock, $s=\SI{0.15}{\metre}$ is the step length, and $R_{\min}=\SI{0.5}{\metre}$ is the minimum turning radius. The resulting maximum heading change was \SI{0.3}{\radian} per step. Setting $R_{\min}$ above the QCar's geometric minimum of \SI{0.44}{\metre} (Section~\ref{sec:qcar}) ensured that the motion primitives respected its curvature limit. Expanded states were stored in a closed set discretised by costmap cell and 72 heading bins of \SI{5}{\degree} each.

The accumulated cost $g$ combined path length, a turning penalty (weight 1.2), and normalised costmap cost (weight 2.0), favouring paths away from obstacles and towards the corridor centreline. The heuristic $h$ used an 8-connected Dijkstra cost-to-go from the goal, incorporating the same costmap term to account for corridor geometry. Each motion primitive was collision-checked along its length at intervals of half the costmap resolution, with unknown and inscribed cells treated as occupied. Search terminated within \SI{0.2}{\metre} of the goal position or after 2000 expansions, without enforcing goal heading.

The resulting path was smoothed iteratively while preserving its endpoints. Each interior point $\mathbf{x}_i$ was adjusted towards its original position $\mathbf{x}_i^0$ and the midpoint of its neighbours:
\begin{equation}\label{eq:smoother}
\mathbf{x}_i \leftarrow \mathbf{x}_i + w_d\left(\mathbf{x}_i^0-\mathbf{x}_i\right) + w_s\left(\mathbf{x}_{i-1}+\mathbf{x}_{i+1}-2\mathbf{x}_i\right)
\end{equation}
where $w_d=0.5$ and $w_s=0.3$. Smoothing continued for up to 50 iterations or until the total displacement per iteration fell below \SI{0.05}{\metre}. Updates were accepted only when the new point and its connecting segment to the preceding point were collision-free. The complete smoothed path was not subsequently revalidated, and path headings were recomputed from its final geometry.

\subsection{Regulated Pure Pursuit}
\label{sec:rpp}

\begin{figure*}[!tbp]
\centering
\begin{tikzpicture}[
    font=\sffamily\fontsize{8}{9.6}\selectfont,
    >={Stealth[length=3.5pt]},
    x=9mm, y=9mm,
    path/.style={black!45, line width=1.6pt},
    car/.style={draw=black, fill=black!10, line width=0.5pt},
    pt/.style={circle, fill=black, inner sep=0pt, minimum size=3pt},
    dim/.style={<->, black, line width=0.4pt},
    guide/.style={black!60, dashed, line width=0.4pt},
    panel/.style={font=\sffamily\bfseries\fontsize{8}{9.6}\selectfont, anchor=north west},
]
\begin{scope}
\node[panel] at (-0.8,2.85) {(a) RPP};
\draw[path] plot[smooth, tension=0.7] coordinates {(-0.6,1.0) (0.8,1.05) (2.0,1.25) (2.8,1.9)};
\draw[car] (-0.3,-0.13) rectangle (0.3,0.13);
\node[pt] at (0,0) {};
\coordinate (P) at (2.0,1.25);
\coordinate (C) at (0,2.225);
\draw[guide] (0,0) -- node[above, sloped] {$L_d$} (P);
\draw[black, line width=0.7pt] (0,0) arc[start angle=-90, end angle=-25.99, radius=2.225];
\draw[guide] (C) -- node[above, sloped] {$1/\kappa$} (P);
\node[pt, minimum size=2pt] at (C) {};
\node[pt] at (P) {};
\node[below right=1pt] at (P) {lookahead};
\node at (1.2,-0.55) {$\kappa = 2y/(x^2+y^2)$};
\end{scope}

\begin{scope}[xshift=45mm]
\node[panel] at (-0.8,2.85) {(b) Stanley};
\coordinate (A) at (-0.4,0.9);
\coordinate (B) at (3.0,1.7);
\draw[path] (A) -- (B);
\draw[car] (-0.3,-0.13) rectangle (0.3,0.13);
\node[pt] at (0,0) {};
\node[above=1pt] at (0,0.13) {centre};
\coordinate (F) at (1.4,0);
\draw[black, line width=0.5pt] (0,0) -- (F);
\node[pt] at (F) {};
\node[right=1pt] at (F) {ref.\ point};
\draw[dim] (0,-0.32) -- node[below] {0.256\,m} (1.4,-0.32);
\coordinate (N) at ($(A)!(F)!(B)$);
\draw[dim] (F) -- node[right=1pt] {$e$} (N);
\node[pt] at (N) {};
\node[above left=1pt] at (N) {nearest pose};
\draw[guide] (N) -- ++(1.1,0);
\draw[->, line width=0.4pt] ([shift={(0:0.8)}]N) arc[start angle=0, end angle=13.24, radius=0.8];
\node at ([shift={(4:1.3)}]N) {$\psi$};
\end{scope}

\begin{scope}[xshift=90mm]
\node[panel] at (-0.8,2.85) {(c) Vector Pursuit};
\draw[path] plot[smooth, tension=0.7] coordinates {(-0.6,0.9) (0.3,0.95) (1.1,1.15) (1.8,1.1) (2.5,0.75) (2.9,0.35)};
\draw[car] (-0.3,-0.13) rectangle (0.3,0.13);
\node[pt] at (0,0) {};
\coordinate (P) at (2.5,0.75);
\draw[guide] (0,0) -- (1.2,0);
\draw[black, line width=0.5pt] (0,0) -- node[below, sloped, pos=0.8] {$l_d$} (P);
\node[pt] at (P) {};
\node[above right=1pt, align=left] at (P) {look-\\ahead};
\draw[->, line width=0.4pt] (0.9,0) arc[start angle=0, end angle=16.70, radius=0.9];
\node at (1.1,0.13) {$\alpha$};
\coordinate (N0) at (0.3,0.95);
\coordinate (N1) at (1.1,1.15);
\draw[black, line width=1pt] (N0) -- (N1);
\node[pt] at (N0) {};
\node[pt] at (N1) {};
\node[below left=1pt, align=right] at (N0) {nearest\\pose};
\node[above=2pt, align=center] at ($(N0)!0.5!(N1)$) {chord\\${\geq}\,0.25$\,m};
\draw[guide] (N0) -- ++(1.0,0);
\draw[->, line width=0.4pt] ([shift={(0:0.75)}]N0) arc[start angle=0, end angle=14.04, radius=0.75];
\node[anchor=north west] at (1.05,0.93) {$\delta_{\mathrm{orient}}$};
\node at (1.2,-0.55) {$\delta_{\mathrm{geom}} = \arctan(2L\sin\alpha/l_d)$};
\end{scope}
\end{tikzpicture}
\caption{Geometry of the three controllers as implemented}\label{fig:controllers}
\end{figure*}
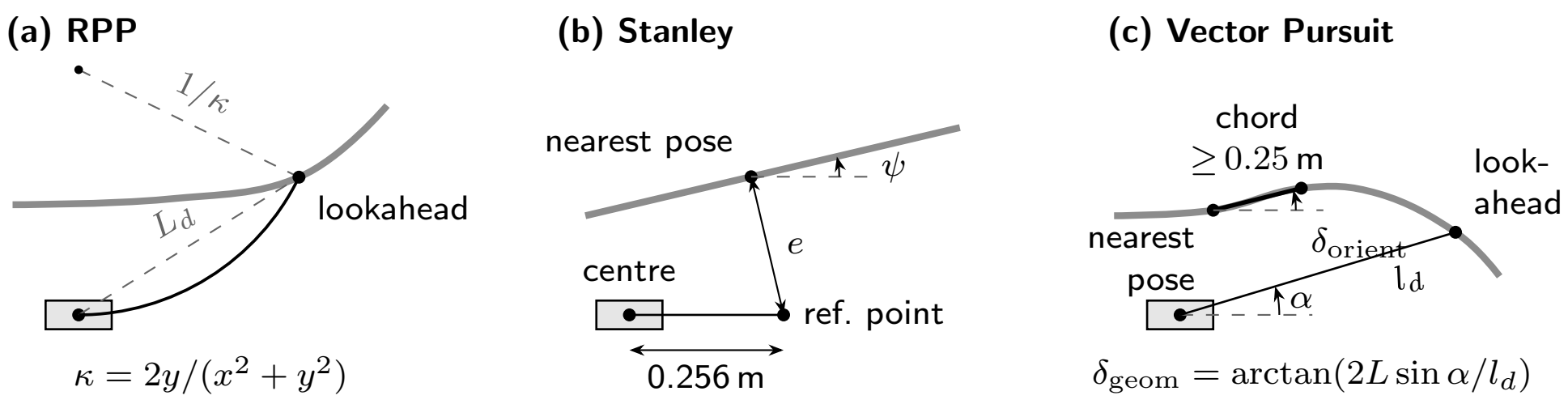

Fig.~\ref{fig:controllers} illustrates the geometry of the three implemented controllers.

The first controller was based on pure pursuit \cite{coulter1992purepursuit}, incorporating curvature-dependent speed regulation adapted from RPP \cite{macenski2023regulatedpurepursuit}. At each control cycle, the path was pruned to the closest pose while preserving forward progress, preventing the lookahead point from returning to previously traversed sections. This was particularly relevant to the hook-shaped track, where separate route segments passed close together. The lookahead point was selected as the first remaining path pose at least $L_d=\SI{0.6}{\metre}$ from the vehicle. Its coordinates $(x,y)$ in the vehicle frame determined the commanded curvature:
\begin{equation}\label{eq:rpp_curv}
\kappa = \frac{2y}{x^2+y^2}
\end{equation}

The commanded speed $v$ was scaled by 0.75 for $|\kappa|>\SI{0.3}{\per\metre}$ and by 0.5 for $|\kappa|>\SI{0.5}{\per\metre}$. Angular velocity $\omega=v\kappa$ was limited to \SI{1.0}{\radian\per\second} and converted to a steering angle using the bicycle model:
\begin{equation}\label{eq:bicycle}
\delta = \arctan\left(\frac{L\omega}{v}\right),
\end{equation}
with steering constrained to the vehicle's configured limits. The controller used neither cross-track nor heading-error feedback, relying instead on lookahead geometry, which attenuated path variations shorter than the lookahead distance. Unlike the full RPP formulation, proximity-based speed reduction, approach-velocity scaling, and collision checking were not implemented.

\subsection{Stanley}
\label{sec:stanley}

The second controller implemented the kinematic Stanley steering law \cite{thrun2006stanley,hoffmann2007stanley}, combining heading alignment with cross-track error correction. The reference point was positioned \SI{0.256}{\metre} ahead of the vehicle centre, approximately \SI{0.13}{\metre} ahead of the front axle. At each cycle, the path was pruned as in RPP, and the pose nearest to this reference point was selected. Cross-track error $e$ was defined as the lateral offset of the selected pose in the vehicle frame, while heading error $\psi$ was calculated from the orientation of the outgoing path segment. The steering command was
\begin{equation}\label{eq:stanley}
\delta = \psi + \arctan\left(\frac{k e}{\max\left(|v|, v_0\right)}\right),
\qquad v_0 = \SI{0.1}{\metre\per\second}
\end{equation}
with $k=0.80$. Although the denominator includes a minimum-speed threshold $v_0$, the controller server supplied zero as the current velocity in every cycle. Consequently, the denominator remained fixed at $v_0$, yielding an effective cross-track gain of $k/v_0=\SI{8}{\per\metre}$ independent of actual vehicle speed. The steering command was limited to $\pm\SI{0.5}{\radian}$.

For consistency with RPP, the same curvature-dependent speed regulation was applied, using the curvature implied by the steering command, $\kappa=|\tan\delta|/L$. Unlike RPP, Stanley corrected errors at the nearest path pose rather than using a lookahead point, making its response more sensitive to local path variations. The yaw-rate and steering-delay damping terms of the full Stanley formulation were not implemented.

\subsection{Vector Pursuit}
\label{sec:vp}

The third controller was a geometric path tracker inspired by Vector Pursuit \cite{wit2000vectorpursuit}, combining position and orientation errors without implementing the original screw-theoretic formulation. Its steering command was a weighted sum of geometric and orientation terms:
\begin{equation}\label{eq:vp}
\begin{aligned}
\delta &= k_t\,\delta_{\mathrm{geom}} + k_r\,\delta_{\mathrm{orient}},\\
\delta_{\mathrm{geom}} &= \arctan\left(\frac{2L\sin\alpha}{l_d}\right)
\end{aligned}
\end{equation}
where $\delta_{\mathrm{geom}}$ was the pure pursuit steering angle towards the first path pose at least \SI{0.6}{\metre} from the vehicle centre, with bearing $\alpha$ and distance $l_d$ ($l_d \geq \SI{0.6}{\metre}$) expressed in the vehicle frame. The orientation term $\delta_{\mathrm{orient}}$ was the heading of the chord connecting the nearest path pose to a pose at least \SI{0.25}{\metre} further along the path, expressed in the vehicle frame. Path pruning and lookahead selection followed RPP. The weights were $k_t=1.5$ and $k_r=2.0$, with the resulting steering command limited to $\pm\SI{0.5}{\radian}$.

Speed was scaled by 0.75 for $|\delta|>\SI{0.30}{\radian}$ and by 0.5 for $|\delta|>\SI{0.425}{\radian}$. Unlike RPP and Stanley, this regulation used steering-angle rather than curvature thresholds.

The non-normalised weights and chord-based heading term prevent exact curvature tracking. For a vehicle on a circular path of radius $R$, the steering command is approximately $\delta \approx (k_t L + k_r c/2)/R$, where $c$ is the chord length. This is approximately 2.5 times the steady-state requirement $L/R$, requiring an inward lateral offset to maintain the curve. The controller therefore combines RPP's lookahead-based steering with a local heading correction similar to Stanley.

\subsection{Controller tuning protocol}
\label{sec:tuning}

Controller parameters were tuned before benchmarking using a separate track containing a sharp U-turn, rather than the evaluation route. Tuning was conducted in simulation and on hardware, prioritising collision-free route completion, command smoothness, and speed. The principal adjusted parameters were commanded linear velocity and Stanley's cross-track gain $k$. Adjustments were made iteratively until further changes increased cross-track error.

Structural parameters were shared where applicable, including the \SI{0.6}{\metre} lookahead distance for RPP and Vector Pursuit, the $\pm\SI{0.5}{\radian}$ steering limit, and the two-stage speed-regulation approach. Controller-specific parameters were then fixed across both planners and environments.

During benchmarking, all planner--controller combinations used the same nominal speed within each environment: \SI{0.35}{\metre\per\second} on hardware and \SI{0.30}{\metre\per\second} in simulation. These common setpoints avoided introducing commanded-speed differences between configurations within each environment. Parameters were not retuned for individual planners, preserving the planner--controller interactions under investigation. Table~\ref{tab:params} summarises the final settings. Stanley's sensitivity to its cross-track gain was evaluated separately in simulation (Section~\ref{sec:gain}).

\begin{table}[!tbp]
\caption{Planner and controller parameters used in the benchmark. Environment-specific values are indicated}
\label{tab:params}
\begin{tabular}{@{}l>{\raggedright\arraybackslash}p{0.5\linewidth}@{}}
\toprule
Parameter & Value \\
\midrule

\multicolumn{2}{@{}l}{\textit{Hybrid A*}} \\
\quad Step length & \SI{0.15}{\metre} \\
\quad Minimum turning radius $R_{\min}$ & \SI{0.5}{\metre} \\
\quad Steering inputs & $\{-1,-0.5,0,0.5,1\}$ (fraction of full lock) \\
\quad Heading bins & 72 \\
\quad Cost weights & Turning: 1.2; costmap: 2.0 \\
\quad Search limits & Goal tolerance: \SI{0.2}{\metre}; maximum expansions: 2000 \\
\quad Smoother & $w_d=0.5$, $w_s=0.3$; maximum 50 iterations; displacement threshold \SI{0.05}{\metre} \\

\midrule
\multicolumn{2}{@{}l}{\textit{NavFn}} \\
\quad Search method & Dijkstra expansion \\
\quad Goal tolerance & \SI{0.15}{\metre} \\
\quad Unknown-space planning & Enabled \\

\midrule
\multicolumn{2}{@{}l}{\textit{Regulated Pure Pursuit}} \\
\quad Lookahead distance & \SI{0.6}{\metre} (fixed) \\
\quad Speed regulation & $\times 0.75$ for $|\kappa|>\SI{0.3}{\per\metre}$; $\times 0.5$ for $|\kappa|>\SI{0.5}{\per\metre}$ \\
\quad Angular-rate limit & \SI{1.0}{\radian\per\second} \\

\midrule
\multicolumn{2}{@{}l}{\textit{Stanley}} \\
\quad Cross-track gain $k$ & 0.80 \\
\quad Minimum speed $v_0$ & \SI{0.1}{\metre\per\second} \\
\quad Reference-point offset & \SI{0.256}{\metre} ahead of vehicle centre \\
\quad Speed regulation & RPP thresholds applied to $\kappa=|\tan\delta|/L$ \\

\midrule
\multicolumn{2}{@{}l}{\textit{Vector Pursuit}} \\
\quad Lookahead distance & \SI{0.6}{\metre} \\
\quad Weights $k_t$, $k_r$ & 1.5, 2.0 \\
\quad Heading chord & At least \SI{0.25}{\metre} \\
\quad Speed regulation & $\times 0.75$ for $|\delta|>\SI{0.30}{\radian}$; $\times 0.5$ for $|\delta|>\SI{0.425}{\radian}$ \\

\midrule
\multicolumn{2}{@{}l}{\textit{Shared settings}} \\
\quad Steering limit & $\pm\SI{0.5}{\radian}$ \\
\quad Controller frequency & \SI{20}{\hertz} \\
\quad Commanded speed & Hardware: \SI{0.35}{\metre\per\second}; simulation: \SI{0.30}{\metre\per\second} \\
\quad Goal check & Position within \SI{0.15}{\metre} of final plan pose; heading not checked \\

\midrule
\multicolumn{2}{@{}l}{\textit{Costmaps (Hybrid A*, NavFn, and controllers)}} \\
\quad Resolution & \SI{0.05}{\metre} \\
\quad Robot radius & \SI{0.24}{\metre} \\
\quad Inflation radius & \SI{0.55}{\metre} \\

\bottomrule
\end{tabular}
\end{table}
  \section{Experimental Design}\label{sec:expdesign}
\subsection{Experimental design and run protocol}
\label{sec:protocol}

The experiment followed a full factorial design comprising three controllers, two planners, and two environments (simulation and hardware), with five valid runs per combination, yielding 60 benchmark runs. Each run followed the hook-shaped route described in Section~\ref{sec:track}, using the single-plan navigation procedure established in Section~\ref{sec:nav}.

For hardware trials, the vehicle was manually positioned at the start before each run, with standard deviations of \SI{0.03}{\metre} in position and \SI{1.7}{\degree} in heading across runs. Runs lasted \SIrange{15}{27}{\second} from first motion to goal arrival, with all hardware trials conducted within a fixed battery operating range. In simulation, the vehicle was reset to an identical start pose before each run. Benchmark speeds were held constant across configurations within each environment (Section~\ref{sec:tuning}). Runs with non-benchmark speed settings, degraded motion-capture data, or incomplete recordings were repeated.

Each run recorded vehicle pose, the global plan, and velocity commands. Controller execution timing and system load were additionally recorded on hardware. All performance metrics were calculated offline from these recordings (Section~\ref{sec:kpi}).

Stanley's cross-track gain sensitivity was evaluated separately in simulation. Both planners were tested at $k \in \{0.40, 0.54, 0.60, 0.70\}$, with three runs per value and all other parameters unchanged. The benchmark gain of $k=0.80$ was represented by the main simulation runs.

\subsection{Performance and computational metrics}
\label{sec:kpi}

All KPIs were calculated offline using a common script across both environments. Each run was evaluated against its recorded global plan, using ground-truth pose expressed in the map frame: motion capture on hardware and simulator pose in simulation. The evaluation window extended from the first to the last velocity command with linear speed exceeding \SI{0.01}{\metre\per\second}. Simulation timestamps followed the simulation clock.

\paragraph{Tracking accuracy.}
Cross-track error (CTE) $e_i$ was defined as the shortest distance from each measured pose to the nearest plan segment, with the projection constrained to that segment. Tracking accuracy was summarised by the root-mean-square error (RMSE) and maximum deviation:
\begin{equation}\label{eq:cte}
\mathrm{CTE}_{\mathrm{RMSE}} = \sqrt{\frac{1}{N}\sum_{i=1}^{N} e_i^2},
\qquad
\mathrm{CTE}_{\max} = \max_i e_i
\end{equation}
Heading error was calculated as the difference between vehicle yaw and the orientation of the nearest plan segment, wrapped to $\pm\pi$ and reported as an RMS value in degrees.

\paragraph{Command smoothness.}
Commanded linear and angular velocities were resampled at \SI{20}{\hertz} using a zero-order hold. Their rates of change were approximated by finite differences and summarised as RMS linear and angular accelerations:
\begin{equation}\label{eq:acc}
a_{\mathrm{RMS}} = \sqrt{\frac{1}{M-1}\sum_{i=1}^{M-1}\left(\frac{u_{i+1} - u_i}{\Delta t}\right)^{2}},
\qquad
\Delta t = \SI{0.05}{\second}
\end{equation}
where $u_i$ denotes the corresponding linear or angular velocity command and $M$ the number of resampled commands. The evaluation window excluded the initial transition from rest and the final command to stop. These metrics quantify variations in controller commands rather than measured vehicle accelerations, avoiding direct dependence on the physical actuation response.

\paragraph{Efficiency.}
Mission time was defined as the duration of the evaluation window. Driven distance was calculated by interpolating the pose trajectory to \SI{10}{\hertz} and summing successive displacements. Average speed was the driven distance divided by mission time. Path efficiency was defined as the ratio of driven to planned distance, with values above unity indicating additional travel. Planned distance was calculated after thinning successive plan points to a minimum spacing of \SI{0.1}{\metre}, preventing grid-scale variations in NavFn paths from inflating their lengths. Final goal error was the distance between the last recorded pose and the plan endpoint.

\paragraph{Computational performance.}
On hardware, controller execution time and control-loop period were recorded for each cycle. Execution time was summarised by its mean, 95th percentile, and maximum; loop period by its mean, standard deviation, and 95th percentile. Planning time was recorded for each global plan. Total CPU utilisation across all cores, per-process CPU utilisation, and memory usage were sampled at \SI{2}{\hertz}. Computational metrics covered the interval from the first planning request to the final motion command.
\subsection{Statistical analysis}
\label{sec:stats}

Individual runs were the unit of analysis, with five counted runs per environment--controller--planner combination (Section~\ref{sec:protocol}). Computational performance metrics were unavailable for some hardware runs because the corresponding topics were not logged in the recorded rosbags; these metrics were therefore summarised from the remaining valid recordings, with the corresponding sample size $n$ reported. Each combination was summarised by its mean and standard deviation (SD).

Simulation and hardware were analysed separately because their longitudinal responses differed (Section~\ref{sec:sim}); data were neither pooled nor compared using absolute KPI values. Within each environment, three effects were estimated: the planner effect, defined as the Hybrid A*--NavFn difference for each controller; controller effects, defined as pairwise controller differences within each planner; and planner--controller interactions, defined as differences in planner effects between controllers. The interaction therefore indicates whether planner influence depends on controller selection.

Effects were expressed as mean differences in the original KPI units, with 95\% confidence intervals obtained through bootstrap resampling \cite{efron1993bootstrap}. Runs were resampled with replacement within each combination for \num{10000} iterations, and interval bounds were taken from the 2.5th and 97.5th percentiles of the resulting differences. An effect was considered resolved when its interval excluded zero. Significance tests and standardised effect sizes were not used because near-deterministic simulation produced extremely small within-cell variation, occasionally below a millimetre, while the five hardware runs per combination provided limited statistical power.

Simulation--hardware agreement was evaluated separately for each KPI using the Spearman rank correlation across the six configuration means \cite{spearman1904proof}, agreement in controller rankings within each planner, and agreement in the sign of each controller's planner effect. These measures assessed the transfer of relative performance patterns rather than absolute KPI values.

Computational metrics were summarised descriptively from hardware runs using Hybrid A*, the more computationally demanding planner. Available sample sizes were $n=3$, $4$, and $5$ for RPP, Stanley, and Vector Pursuit, respectively. Planning times were pooled across controllers because planner execution preceded and was independent of controller operation. \section{Results}\label{sec:results}

\subsection{Overall performance}
\label{sec:overall}

\begin{figure*}[!tbp]
\centering
\includegraphics[width=\textwidth]{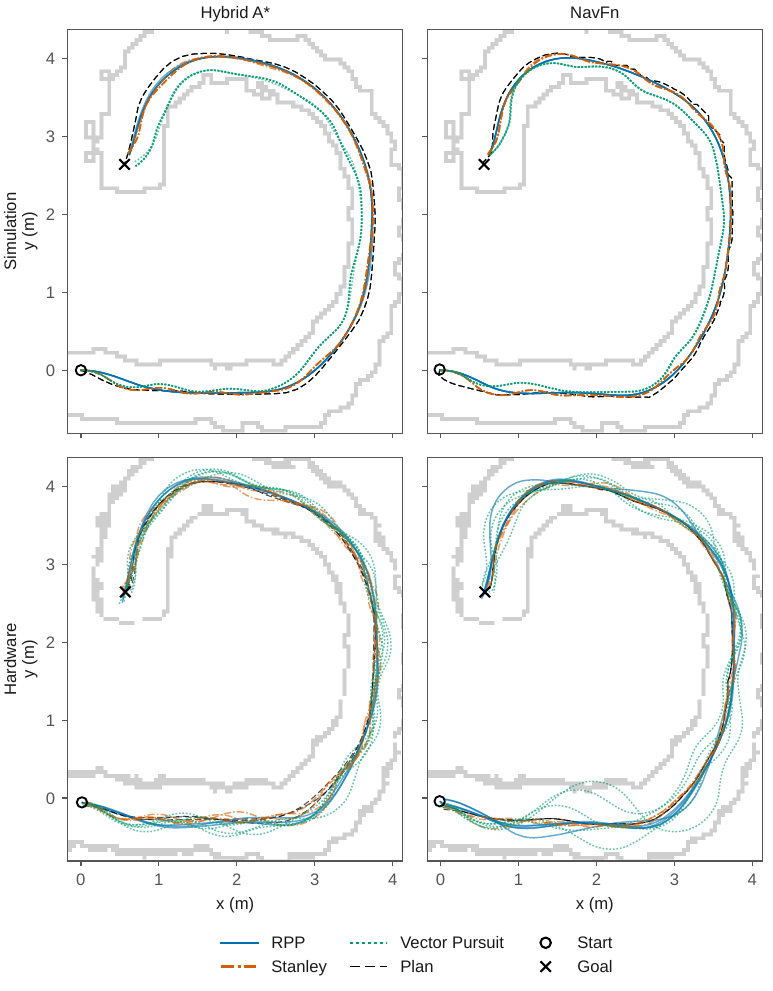}
\caption{Driven trajectories of all counted runs, by environment (rows) and planner (columns). Controllers are distinguished by colour and line style. Dashed black lines: plans. $\circ$: start; $\times$: goal. Grey: navigation map}\label{fig:trajectories}
\end{figure*}

All 60 counted runs reached the goal. Table~\ref{tab:summary}
reports the five headline KPIs for each planner--controller
combination, while Fig.~\ref{fig:trajectories} shows the recorded
trajectories. Supplementary KPI results are provided in
Table~\ref{tab:all_kpis} in the Appendix.

\begin{table}[!tbp]
\footnotesize
\setlength{\tabcolsep}{4pt}
\newcommand{\kpicell}[2]{\begin{tabular}[t]{@{}c@{}}$#1$\\[-1pt]$\pm #2$\end{tabular}}
\caption{Headline performance metrics for each planner--controller combination (mean $\pm$ SD, $n=5$)}
\label{tab:summary}

\begin{tabular}{@{}llccccc@{}}
\toprule
Controller & Planner
& \begin{tabular}[b]{@{}c@{}}CTE\\RMSE\\(\si{\metre})\end{tabular}
& \begin{tabular}[b]{@{}c@{}}Max.\\CTE\\(\si{\metre})\end{tabular}
& \begin{tabular}[b]{@{}c@{}}Linear\\accel.\ RMS\\(\si{\metre\per\second\squared})\end{tabular}
& \begin{tabular}[b]{@{}c@{}}Angular\\accel.\ RMS\\(\si{\radian\per\second\squared})\end{tabular}
& \begin{tabular}[b]{@{}c@{}}Mission\\time\\(\si{\second})\end{tabular} \\
\midrule

\multicolumn{7}{@{}l}{\textit{Simulation}} \\
RPP & Hybrid A* & \kpicell{0.0476}{0.000173} & \kpicell{0.109}{0.000179} & \kpicell{0.213}{0.00775} & \kpicell{0.121}{0.00219} & \kpicell{55.4}{0.472} \\[2pt]
& NavFn & \kpicell{0.0546}{0.0000548} & \kpicell{0.156}{0} & \kpicell{0.293}{0.00727} & \kpicell{0.137}{0.00121} & \kpicell{53.7}{0.0821} \\[2pt]
Stanley & Hybrid A* & \kpicell{0.0484}{0.000295} & \kpicell{0.0962}{0.00174} & \kpicell{0.718}{0.0186} & \kpicell{0.867}{0.0228} & \kpicell{56.4}{0.685} \\[2pt]
& NavFn & \kpicell{0.0472}{0.000497} & \kpicell{0.145}{0} & \kpicell{0.538}{0.0210} & \kpicell{0.890}{0.0190} & \kpicell{65.2}{0.129} \\[2pt]
Vector Pursuit & Hybrid A* & \kpicell{0.184}{0.00160} & \kpicell{0.328}{0.0102} & \kpicell{0.542}{0.0350} & \kpicell{1.09}{0.0344} & \kpicell{44.2}{0.932} \\[2pt]
& NavFn & \kpicell{0.124}{0.00129} & \kpicell{0.189}{0.000858} & \kpicell{0.441}{0.0129} & \kpicell{1.06}{0.0279} & \kpicell{47.4}{0.288} \\[2pt]

\midrule
\multicolumn{7}{@{}l}{\textit{Hardware}} \\
RPP & Hybrid A* & \kpicell{0.0565}{0.00443} & \kpicell{0.121}{0.0108} & \kpicell{0.530}{0.0167} & \kpicell{0.377}{0.0112} & \kpicell{15.8}{0.357} \\[2pt]
& NavFn & \kpicell{0.0698}{0.0330} & \kpicell{0.136}{0.0642} & \kpicell{0.397}{0.0169} & \kpicell{0.236}{0.0209} & \kpicell{21.8}{2.88} \\[2pt]
Stanley & Hybrid A* & \kpicell{0.0454}{0.0141} & \kpicell{0.0909}{0.0301} & \kpicell{0.986}{0.0710} & \kpicell{1.41}{0.0922} & \kpicell{23.9}{1.28} \\[2pt]
& NavFn & \kpicell{0.0343}{0.00892} & \kpicell{0.0935}{0.0356} & \kpicell{0.787}{0.0300} & \kpicell{0.882}{0.0802} & \kpicell{26.5}{0.595} \\[2pt]
Vector Pursuit & Hybrid A* & \kpicell{0.0939}{0.0233} & \kpicell{0.191}{0.0444} & \kpicell{0.878}{0.0554} & \kpicell{2.13}{0.206} & \kpicell{17.4}{1.08} \\[2pt]
& NavFn & \kpicell{0.144}{0.0673} & \kpicell{0.327}{0.166} & \kpicell{0.603}{0.0808} & \kpicell{1.29}{0.237} & \kpicell{22.8}{3.36} \\[2pt]

\bottomrule
\end{tabular}

\par\smallskip
{\footnotesize\emph{Note:} Simulation mission times use the simulation clock. Absolute KPI values are not compared between environments (Section~\ref{sec:stats}).\par}

\end{table} 

Vector Pursuit recorded the highest cross-track error in every configuration. Its CTE RMSE ranged from \SIrange{0.124}{0.184}{\metre} in simulation and from \SIrange{0.0939}{0.144}{\metre} on hardware, whereas all RPP and Stanley configurations remained below \SI{0.070}{\metre}. With Hybrid A* in simulation, Vector Pursuit maintained an inward offset along the main bend. On hardware with NavFn, it exhibited substantial lateral excursions along the initial straight, producing a mean maximum CTE of \SI{0.327}{\metre}, with a maximum of \SI{0.55}{\metre} in one run.

RPP and Stanley achieved CTE RMSE values below \SI{0.055}{\metre} in simulation. With Hybrid A*, their errors were \SI{0.0476}{\metre} and \SI{0.0484}{\metre}, respectively; with NavFn, they were \SI{0.0546}{\metre} and \SI{0.0472}{\metre}. On hardware, Stanley achieved lower CTE RMSE than RPP with both planners: \SI{0.0454}{\metre} versus \SI{0.0565}{\metre} with Hybrid A*, and \SI{0.0343}{\metre} versus \SI{0.0698}{\metre} with NavFn.

Angular acceleration RMS consistently ranked the controllers RPP $<$ Stanley $<$ Vector Pursuit across both planners and environments. In simulation, the respective ranges were \SIrange{0.121}{0.137}{\radian\per\second\squared}, \SIrange{0.867}{0.890}{\radian\per\second\squared}, and \SIrange{1.06}{1.09}{\radian\per\second\squared}. On hardware, they were \SIrange{0.236}{0.377}{\radian\per\second\squared}, \SIrange{0.882}{1.41}{\radian\per\second\squared}, and \SIrange{1.29}{2.13}{\radian\per\second\squared}, respectively. Linear acceleration RMS followed the ordering RPP $<$ Vector Pursuit $<$ Stanley in every configuration.

Stanley recorded the longest mission time in all four environment--planner combinations, exceeding the other controllers by \SIrange{1.0}{17.8}{\second}. Mission times were evaluated within each environment, as specified in Section~\ref{sec:stats}.
\subsection{Planner--controller interaction}
\label{sec:interaction}

\begin{figure*}[!tbp]
\centering
\includegraphics{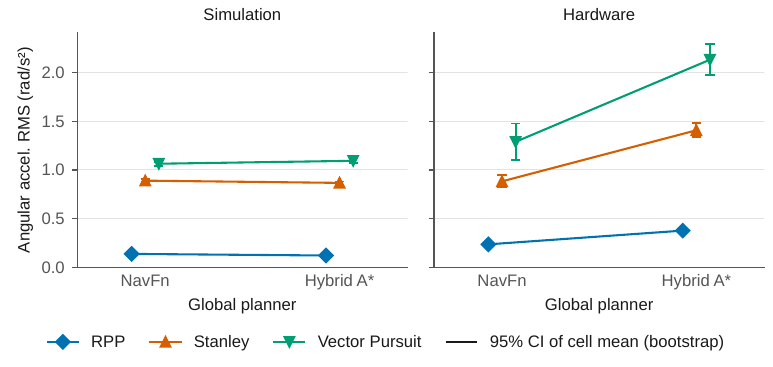}
\caption{Angular acceleration RMS of each controller with each global planner, in simulation and on hardware. Points: cell means ($n = 5$); error bars: 95\% bootstrap confidence intervals of the cell mean. The controllers are offset horizontally for legibility}\label{fig:interaction}
\end{figure*}

Fig.~\ref{fig:interaction} shows angular acceleration RMS for each controller with each planner, and Table~\ref{tab:planner_effects} gives the planner effect (Hybrid A* minus NavFn) on linear acceleration RMS and CTE RMSE, both with 95\% bootstrap confidence intervals (Section~\ref{sec:stats}).

In simulation, planner selection had a limited effect on angular acceleration RMS. Hybrid A* reduced the metric by \SI{0.016}{\radian\per\second\squared} for RPP and \SI{0.023}{\radian\per\second\squared} for Stanley, while increasing it by \SI{0.030}{\radian\per\second\squared} for Vector Pursuit. The corresponding effects on linear acceleration RMS were $-0.080$, $+0.179$, and $+0.101$\,\si{\metre\per\second\squared}, respectively.

On hardware, Hybrid A* increased both command-acceleration RMS metrics for all three controllers. Angular acceleration RMS increased by \SI{0.141}{\radian\per\second\squared} for RPP, \SI{0.524}{\radian\per\second\squared} for Stanley, and \SI{0.843}{\radian\per\second\squared} for Vector Pursuit. Linear acceleration RMS increased by \SI{0.133}{\metre\per\second\squared}, \SI{0.200}{\metre\per\second\squared}, and \SI{0.276}{\metre\per\second\squared}, respectively, with every confidence interval excluding zero (Table~\ref{tab:planner_effects}). The differences between the controllers' angular-acceleration planner effects were \SI{0.38}{\radian\per\second\squared} (Stanley--RPP), \SI{0.70}{\radian\per\second\squared} (Vector Pursuit--RPP), and \SI{0.32}{\radian\per\second\squared} (Vector Pursuit--Stanley) on hardware, compared with no more than \SI{0.053}{\radian\per\second\squared} in simulation.

Planner effects on tracking accuracy were less consistent. In simulation, Hybrid A* changed CTE RMSE by $-7$\,\si{\milli\metre} for RPP, $+1$\,\si{\milli\metre} for Stanley, and $+60$\,\si{\milli\metre} for Vector Pursuit. No planner effect on CTE RMSE was resolved on hardware. For Vector Pursuit, the effect reversed to $-50$\,\si{\milli\metre}, but its confidence interval included zero.

\begin{table}[!htbp]
\footnotesize
\setlength{\tabcolsep}{5pt}
\newcommand{\effcell}[2]{\begin{tabular}[t]{@{}c@{}}#1\\[-1pt][#2]\end{tabular}}
\caption{Planner effect (Hybrid A* $-$ NavFn) on linear acceleration RMS and CTE RMSE: mean difference and 95\% bootstrap confidence interval per controller and environment}
\label{tab:planner_effects}
\begin{tabular}{@{}lcccc@{}}
\toprule
 & \multicolumn{2}{c}{Linear accel.\ RMS (\si{\metre\per\second\squared})} & \multicolumn{2}{c}{CTE RMSE (\si{\milli\metre})} \\
\cmidrule(lr){2-3}\cmidrule(l){4-5}
Controller & Simulation & Hardware & Simulation & Hardware \\
\midrule
RPP & \effcell{$-$0.080}{$-$0.088, $-$0.071} & \effcell{$+$0.133}{0.114, 0.151} & \effcell{$-$7.0}{$-$7.2, $-$6.9} & \effcell{$-$13.3}{$-$41.1, 10.5} \\[2pt]
Stanley & \effcell{$+$0.179}{0.157, 0.201} & \effcell{$+$0.200}{0.136, 0.257} & \effcell{$+$1.2}{0.8, 1.7} & \effcell{$+$11.1}{$-$1.4, 24.3} \\[2pt]
Vector Pursuit & \effcell{$+$0.101}{0.068, 0.126} & \effcell{$+$0.276}{0.201, 0.353} & \effcell{$+$60.4}{58.9, 62.1} & \effcell{$-$49.6}{$-$107.4, 3.1} \\[2pt]
\bottomrule
\end{tabular}
\par\smallskip
{\footnotesize\emph{Note:} $n = 5$ runs per cell. Bootstrap: \num{10000} resamples of runs within each cell (Section~\ref{sec:stats}).\par}
\end{table}

On hardware, all three controllers completed the route faster with Hybrid A*, despite identical commanded speeds across planners. Mission times decreased by \SIrange{2.6}{6.0}{\second}, while average speeds increased by \SIrange{0.05}{0.17}{\metre\per\second}. In simulation, planner-dependent differences in average speed remained below \SI{0.03}{\metre\per\second}.

\subsection{Simulation-to-hardware transfer}
\label{sec:transfer}

Table~\ref{tab:agreement} summarises simulation--hardware agreement in controller rankings and planner effects for each KPI.

\begin{table}[!tbp]
\caption{Simulation--hardware agreement in KPI rankings and planner-effect directions}
\label{tab:agreement}

\begin{tabular}{@{}lcccccc@{}}
\toprule
& & \multicolumn{2}{c}{Controller order agrees}
& \multicolumn{3}{c}{Planner-effect sign agrees} \\
\cmidrule(lr){3-4}
\cmidrule(l){5-7}

KPI & Spearman $\rho$ & Hybrid A* & NavFn & RPP & Stanley & VP \\
\midrule

CTE RMSE
& $0.886$ & \xmark & \cmark & \cmark & \cmark & \xmark \\

Maximum CTE
& $0.886$ & \cmark & \cmark & \cmark & \cmark & \xmark \\

Linear accel.\ RMS
& $0.943$ & \cmark & \cmark & \xmark & \cmark & \cmark \\

Angular accel.\ RMS
& $0.771$ & \cmark & \cmark & \xmark & \xmark & \cmark \\

Mission time
& $0.600$ & \xmark & \xmark & \xmark & \cmark & \cmark \\

Path efficiency
& $-0.714$ & \xmark & \xmark & \cmark & \xmark & \cmark \\

Average speed
& $0.600$
& \xmark & \xmark
& \xmark & \cmark & \cmark \\

Final goal error
& $0.029$
& \xmark & \xmark
& \cmark & \xmark & \xmark \\

Heading error RMS
& $0.371$
& \xmark & \cmark
& \cmark & \xmark & \cmark \\

\bottomrule
\end{tabular}

\par\smallskip
{\footnotesize
\emph{Note:}
\cmark\ indicates agreement and \xmark\ disagreement. VP: Vector Pursuit.
Spearman correlations use the six planner--controller cell means.
Controller order is based on ascending cell means within each planner.
Planner-effect signs are based on
mean(Hybrid A*) $-$ mean(NavFn) for each controller;
sign agreement does not imply a statistically resolved effect.
\par}

\end{table}
Controller rankings for tracking accuracy and command smoothness showed the strongest agreement between environments. Spearman correlations were $\rho=0.89$ for both CTE RMSE and maximum CTE, $0.94$ for linear acceleration RMS, and $0.77$ for angular acceleration RMS. Within each planner, controller rankings were identical across environments for maximum CTE and both acceleration metrics. CTE RMSE rankings were also identical under NavFn; under Hybrid A*, only RPP and Stanley exchanged positions, having differed by \SI{0.8}{\milli\metre} in simulation.

In contrast, planner effects showed limited transfer. For angular acceleration RMS, the effect direction agreed between environments only for Vector Pursuit; for linear acceleration RMS, agreement was observed for Stanley and Vector Pursuit. Planner-dependent changes in angular acceleration RMS did not exceed \SI{0.030}{\radian\per\second\squared} in simulation, whereas Hybrid A* increased the metric by \SIrange{0.141}{0.843}{\radian\per\second\squared} on hardware (Section~\ref{sec:interaction}). Vector Pursuit's CTE RMSE planner effect also reversed, from $+60$\,\si{\milli\metre} in simulation to $-50$\,\si{\milli\metre} on hardware, although the hardware confidence interval included zero.

Agreement was weaker for the remaining KPIs. Mission time and average speed each yielded $\rho=0.60$, with Vector Pursuit completing the route fastest in simulation and RPP on hardware. Stanley was the slowest controller in both environments. Heading error RMS ($\rho=0.37$) and final goal error ($\rho=0.03$) showed weak rank agreement, while path efficiency exhibited a negative correlation ($\rho=-0.71$), indicating substantially different configuration rankings.

Overall, simulation reproduced the relative controller performance in tracking accuracy and command smoothness, but not the magnitude or direction of planner-dependent effects consistently. Agreement in mission time, achieved speed, and secondary KPIs was also weaker.

\subsection{Sensitivity of Stanley to its gain}
\label{sec:gain}

\begin{figure}[!tbp]
\centering
\includegraphics{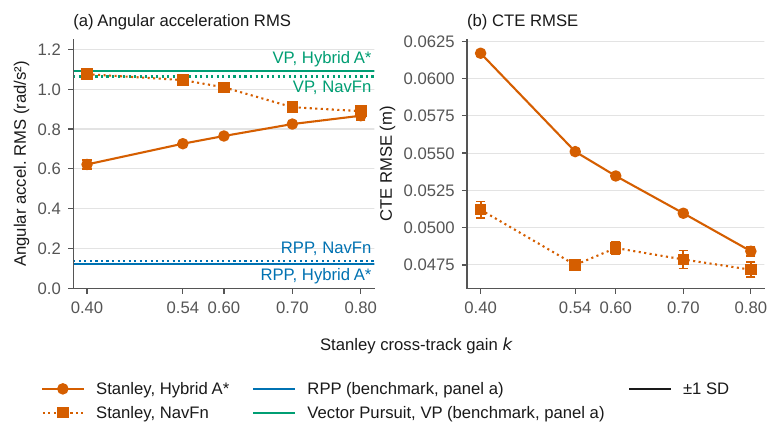}
\caption{Sensitivity of Stanley to its cross-track gain $k$ in simulation. (a)~Angular acceleration RMS against $k$; (b)~CTE RMSE against $k$. Stanley is shown in orange, with Hybrid A* as circles on a solid line and NavFn as squares on a dotted line. The horizontal lines in (a) are the benchmark values of RPP (blue) and Vector Pursuit (VP, green), solid for Hybrid A* and dotted for NavFn. Error bars: $\pm 1$ SD; $n = 3$ runs per point for $k < 0.80$ and $n = 5$ at $k = 0.80$ (the benchmark runs)}\label{fig:gain}
\end{figure}

Stanley's response to its cross-track gain depended on the global planner (Fig.~\ref{fig:gain}). Reducing $k$ from 0.80 to 0.40 with Hybrid A* increased CTE RMSE from \SI{48.4}{\milli\metre} to \SI{61.7}{\milli\metre}, while reducing angular acceleration RMS from \SI{0.867}{\radian\per\second\squared} to \SI{0.622}{\radian\per\second\squared}. With NavFn, the same gain reduction increased both CTE RMSE (\SI{47.2}{\milli\metre} to \SI{51.2}{\milli\metre}) and angular acceleration RMS (\SI{0.890}{\radian\per\second\squared} to \SI{1.08}{\radian\per\second\squared}). The angular acceleration ranking RPP $<$ Stanley $<$ Vector Pursuit remained unchanged across all tested gains with Hybrid A*. With NavFn, Stanley matched or exceeded Vector Pursuit at $k=0.40$ and $k=0.54$, demonstrating that controller smoothness rankings were sensitive to gain selection under this planner.

\subsection{Computational load on the TX2}
\label{sec:compute}

Table~\ref{tab:compute} reports the computational load measured on the Jetson TX2 during Hybrid A* hardware runs ($n=3$, $4$, and $5$ for RPP, Stanley, and Vector Pursuit, respectively). Measurements included the navigation stack, sensor drivers, motion-capture client, and data recording.

\begin{table}[!tbp]
\caption{Computational performance on the Jetson TX2 during Hybrid A* hardware runs}
\label{tab:compute}

\begin{tabular}{@{}lccc@{}}
\toprule
& \begin{tabular}[b]{@{}c@{}}RPP\\($n=3$)\end{tabular}
& \begin{tabular}[b]{@{}c@{}}Stanley\\($n=4$)\end{tabular}
& \begin{tabular}[b]{@{}c@{}}Vector Pursuit\\($n=5$)\end{tabular} \\
\midrule

\multicolumn{4}{@{}l}{\textit{Controller computation time (\si{\milli\second})}} \\
\quad Mean
& $0.420 \pm 0.0556$
& $0.399 \pm 0.0154$
& $0.434 \pm 0.0490$ \\

\quad 95th percentile
& $0.968 \pm 0.189$
& $0.923 \pm 0.160$
& $1.11 \pm 0.249$ \\

\quad Per-run maximum
& $8.26 \pm 3.60$
& $7.40 \pm 4.56$
& $5.40 \pm 1.58$ \\

\midrule
\multicolumn{4}{@{}l}{\textit{Control-loop period (\si{\milli\second})}} \\
\quad Mean
& $50.0 \pm 0.000577$
& $50.0 \pm 0$
& $50.0 \pm 0.000894$ \\

\quad SD
& $0.924 \pm 0.408$
& $0.824 \pm 0.105$
& $1.16 \pm 0.865$ \\

\quad 95th percentile
& $51.0 \pm 0.262$
& $51.0 \pm 0.237$
& $51.0 \pm 0.263$ \\

\midrule
\multicolumn{4}{@{}l}{\textit{System resource utilisation}} \\

\quad Total CPU mean (\si{\percent})
& $63.8 \pm 0.102$
& $63.9 \pm 0.462$
& $63.9 \pm 0.0644$ \\

\quad Total CPU maximum (\si{\percent})
& $65.2 \pm 0.404$
& $65.7 \pm 0.629$
& $65.7 \pm 0.444$ \\

\quad Controller CPU (\si{\percent} of one core)
& $6.07 \pm 0.223$
& $5.96 \pm 0.0566$
& $6.11 \pm 0.159$ \\

\quad Planner CPU (\si{\percent} of one core)
& $4.31 \pm 0.115$
& $3.97 \pm 0.0866$
& $4.16 \pm 0.130$ \\

\quad Peak system RAM (\si{\gibi\byte})
& $4.32 \pm 0.0127$
& $4.33 \pm 0.0126$
& $4.41 \pm 0.0282$ \\

\midrule

Planning time per plan (\si{\milli\second})
& \multicolumn{3}{c}{Mean: 59.8; range: 27.2--123 ($n=12$)} \\

\bottomrule
\end{tabular}

\par\smallskip
{\footnotesize
\emph{Note:}
Values are mean $\pm$ SD across runs of the per-run statistic
specified in each row. Total CPU utilisation is measured across
all online cores; process CPU is expressed as a percentage of
one core. RAM represents system memory in use
(\texttt{MemTotal} $-$ \texttt{MemAvailable}).
Planning time was pooled across all 12 Hybrid A* runs.
Measurements include the navigation stack, sensor drivers,
motion-capture client, and recording processes.
\par}

\end{table}
Mean controller computation times ranged from \SI{0.399}{\milli\second} to \SI{0.434}{\milli\second}, with 95th-percentile values of \SIrange{0.923}{1.11}{\milli\second}. The mean computation times differed by less than \SI{0.04}{\milli\second} across controllers. Mean control-loop periods remained at \SI{50.0}{\milli\second}, with standard deviations of \SIrange{0.824}{1.16}{\milli\second} and 95th-percentile periods of \SI{51.0}{\milli\second}.

Hybrid A* planning required \SIrange{27.2}{123}{\milli\second} per run, with a mean of \SI{59.8}{\milli\second} across 12 runs. Mean total CPU utilisation ranged from \SI{63.8}{\percent} to \SI{63.9}{\percent}, while the controller and planner processes consumed \SIrange{5.96}{6.11}{\percent} and \SIrange{3.97}{4.31}{\percent} of one CPU core, respectively. Mean peak system memory usage ranged from \SI{4.32}{\gibi\byte} to \SI{4.41}{\gibi\byte} of the available \SI{8}{\gibi\byte}. These measurements indicate that controller computation imposed a limited processing burden on the TX2 under the tested workload. \section{Discussion}\label{sec:discussion}

\subsection{Transfer of controller rankings}
\label{sec:disc_transfer}

Simulation reproduced the relative performance of the three controllers in tracking accuracy and command smoothness, with Spearman correlations ranging from 0.77 to 0.94 (Table~\ref{tab:agreement}). Controller rankings were identical across environments for maximum CTE and both command-acceleration metrics. The only discrepancy in CTE RMSE rankings occurred between RPP and Stanley under Hybrid A*, whose simulated errors differed by just \SI{0.8}{\milli\metre}. These results demonstrate that simulation provided reliable controller rankings for lateral tracking accuracy and command smoothness under the tested configurations, despite differences in vehicle dynamics.

This agreement contrasts with the ranking reversals reported in previous studies. Kadian et al. \cite{kadian2020sim2real} observed poor simulation--hardware ranking agreement until simulator parameters were tuned, while Kong et al. \cite{kong2024purepursuitmpc} attributed controller-ranking reversals to computational delays in optimisation-based methods. In the present study, the geometric controllers required less than \SI{0.5}{\milli\second} per command on average, representing less than 1\% of the control period (Section~\ref{sec:compute}). Their low computational demands reduce the relevance of optimisation-related delays observed in earlier comparisons. The rankings transferred despite the simulation's idealised longitudinal response.

In contrast, metrics influenced by longitudinal dynamics showed weaker agreement. Mission time and average speed each yielded $\rho=0.60$, while path efficiency exhibited a negative correlation ($\rho=-0.71$). These discrepancies are consistent with the difference between ideal commanded-speed tracking in simulation and open-loop throttle control on hardware (Section~\ref{sec:sim}). Final goal error showed negligible ranking agreement ($\rho=0.03$), with the hardware results reflecting residual vehicle motion after goal acceptance. Heading error RMS also showed limited agreement ($\rho=0.37$). Thus, simulation reliably ranked the tested controllers for lateral tracking accuracy and command smoothness but did not reproduce their relative performance in mission efficiency or final goal positioning.

\subsection{The planner--controller interaction}
\label{sec:disc_interaction}

The principal simulation--hardware discrepancy concerned planner-dependent command smoothness. In simulation, planner effects on angular acceleration RMS did not exceed \SI{0.030}{\radian\per\second\squared}, and interaction contrasts did not exceed \SI{0.053}{\radian\per\second\squared}. On hardware, Hybrid A* increased angular acceleration RMS by \SIrange{0.141}{0.843}{\radian\per\second\squared}, with interaction contrasts of \SIrange{0.32}{0.70}{\radian\per\second\squared} (Section~\ref{sec:interaction}). Simulation therefore substantially underestimated the influence of planner selection on command smoothness.

On hardware, these effects coincided with differences in achieved vehicle speed. Despite a common nominal speed of \SI{0.35}{\metre\per\second} and a fixed battery operating range, Hybrid A* configurations achieved average speeds \SIrange{0.05}{0.17}{\metre\per\second} higher than their NavFn counterparts. The corresponding differences in simulation remained below \SI{0.03}{\metre\per\second}. The idealised longitudinal response in simulation (Section~\ref{sec:sim}) did not reproduce the speed differences observed with the physical vehicle.

For a fixed curvature profile traversed at constant speed, angular acceleration scales with the square of vehicle speed. Applying this relationship to the measured mean hardware speeds gives predicted Hybrid A*--NavFn angular acceleration ratios of 1.75 for RPP, 1.25 for Stanley, and 1.64 for Vector Pursuit. The observed ratios were 1.60, 1.60, and 1.65, respectively. The agreement for RPP and Vector Pursuit is consistent with achieved speed contributing to their planner-dependent command variations. For Stanley, the observed increase exceeded the speed-based estimate, indicating that speed scaling alone does not account for the difference under these assumptions. Because the planners also generated different paths and vehicle speed varied during execution, these comparisons do not isolate speed as the causal mechanism.

Planner effects on tracking accuracy were less consistent, with no resolved CTE RMSE effects on hardware and a reversal in the observed effect for Vector Pursuit. Taken together, the findings show that simulation reproduced relative controller performance more consistently than planner-dependent effects. For the tested platform, planner selection based on command smoothness therefore required hardware validation.

\subsection{Controller-specific behaviour}
\label{sec:behaviour}

RPP consistently generated the smoothest commands. Its fixed \SI{0.6}{\metre} lookahead and absence of direct cross-track feedback attenuated local path variations (Section~\ref{sec:rpp}), favouring smooth command generation while maintaining competitive tracking accuracy.

Stanley's fixed effective cross-track gain of \SI{8}{\per\metre} made its steering response sensitive to small lateral deviations (Section~\ref{sec:stanley}). At zero heading error, cross-track errors of \SI{1.0}{\centi\metre} and \SI{1.6}{\centi\metre} reached the thresholds at which speed was scaled to 75\% and 50\%, respectively, while \SI{6.8}{\centi\metre} brought steering to its limit. This sensitivity provides an explanation for Stanley's consistently longer mission times despite its tracking accuracy.

The gain-sensitivity experiment further demonstrated the dependence of Stanley's behaviour on planner selection (Section~\ref{sec:gain}). Reducing the cross-track gain improved command smoothness with Hybrid A* but degraded it with NavFn. One possible explanation is the heading-error term's sensitivity to local variations in NavFn's grid-based path, which become relatively more influential as cross-track correction weakens. Consequently, the observed smoothness ranking was stable across the tested gains with Hybrid A*, but not with NavFn.

Vector Pursuit's consistently higher tracking error was consistent with the curvature-tracking bias of the implemented control law (Section~\ref{sec:vp}). Its non-normalised weights and chord-based orientation term required an inward lateral offset during steady turning, consistent with the trajectory observed under Hybrid A* in simulation. The large lateral excursions recorded with NavFn on hardware were not explained by this steady-turn behaviour alone. These findings characterise the geometric implementation evaluated here rather than the original screw-theoretic Vector Pursuit formulation \cite{wit2000vectorpursuit}.

\subsection{Computational load}
\label{sec:disc_compute}

The geometric controllers imposed a limited computational burden on the Jetson TX2, with mean execution times below \SI{0.5}{\milli\second} and differences below \SI{0.04}{\milli\second} between controllers (Section~\ref{sec:compute}). Their execution times were small relative to the \SI{50}{\milli\second} control period, indicating that controller computation was not a limiting factor under the tested workload. Although computational constraints have been identified as a concern for navigation on legacy NVIDIA Jetson platforms \cite{hammami2026sushi}, the planning and control algorithms evaluated here required a small fraction of the available processing resources. This contrasts with the optimisation-based controllers examined by Kong et al. \cite{kong2024purepursuitmpc}, for which computational delays contributed to differences between simulation and hardware performance.

The complete system nevertheless operated under a substantially greater workload, including sensor drivers, motion capture, and data recording. Total CPU utilisation reached \SI{65.7}{\percent} in the reported per-run statistics, while battery capacity constrained the duration of experimental sessions. The computational measurements were limited to Hybrid A* hardware runs, with three to five runs per controller. They therefore establish the feasibility of the tested implementations under the measured conditions, rather than their capacity to accommodate additional computational demands.

\subsection{Scope and limitations}
\label{sec:limitations}

The benchmark comprised five runs per configuration, limiting the precision of hardware estimates, and computational measurements were limited to Hybrid A* hardware runs ($n=3$--$5$ per controller). Experiments were restricted to one vehicle, one track, and one commanded speed per environment. Consequently, the findings do not establish how planner--controller interactions vary with vehicle speed, route geometry, or operating conditions.

Simulation and hardware differed in longitudinal dynamics, with idealised speed tracking in simulation and open-loop throttle control on the QCar. The mechanism underlying the planner-dependent differences in achieved hardware speed was not independently isolated. Although the custom planning and control algorithms were shared, these differences limit the interpretation of simulation--hardware agreement.

Both environments used external ground-truth pose references, excluding the influence of online state-estimation errors. Hardware motion-capture measurements nevertheless incurred uncompensated latency, with the measured retrieval delay corresponding to a positional lag of \SIrange{3}{12}{\milli\metre} (Section~\ref{sec:sim}).

 \section{Conclusion}
\label{sec:conclusion}

This study evaluated three geometric path-tracking controllers (Regulated Pure Pursuit, Stanley, and Vector Pursuit) paired with two global planners (Hybrid A* and NavFn) on a 1/10-scale Ackermann vehicle and in a matched Gazebo simulation. The benchmark comprised 60 runs, with all configurations sharing the custom planning and control algorithms, route geometry, goals, and evaluation pipeline.

Controller rankings for tracking accuracy and command smoothness transferred consistently between environments, with Spearman correlations of 0.77--0.94. However, planner-dependent effects showed substantially weaker agreement. In simulation, planner selection changed angular acceleration RMS by no more than \SI{0.030}{\radian\per\second\squared}, whereas Hybrid A* increased it by \SIrange{0.141}{0.843}{\radian\per\second\squared} on hardware, depending on the controller. These differences coincided with higher achieved speeds under Hybrid A*, which the simulation's idealised longitudinal response did not reproduce. Mission time, average speed, path efficiency, and final goal positioning also showed weaker agreement.

The findings demonstrate that agreement in controller rankings does not guarantee agreement in planner--controller behaviour. Although simulation reproduced the relative tracking accuracy and command smoothness of the tested controllers, it did not consistently predict how individual planner--controller pairings would respond on hardware. Consequently, selecting a controller based on simulation performance alone is insufficient to establish the performance of a specific navigation configuration. Hardware validation remains necessary to assess planner--controller interactions under the intended operating conditions.

All three controllers operated within the Jetson TX2's measured computational budget, with mean command computation times below \SI{0.5}{\milli\second}. Future work will incorporate an optimisation-based controller and extend the benchmark to additional routes, commanded speeds, and experimental repetitions. The navigation stack, simulation environment, and processed KPI results are openly available (see Data availability and Code availability).
\backmatter

\section*{Declarations}
\label{sec:declarations}

\bmhead{Data availability}
The processed KPI results supporting the findings of this study are available in the hardware repository (\url{https://github.com/HWUD-AIMBeyonD/qcar}) and the simulation repository (\url{https://github.com/HWUD-AIMBeyonD/QCarSim}). The raw rosbag recordings are not publicly available due to their size but are available from the corresponding author on reasonable request.

\bmhead{Code availability}
The planner, controller, simulation, and KPI-evaluation source code is available at \url{https://github.com/HWUD-AIMBeyonD/qcar} and \url{https://github.com/HWUD-AIMBeyonD/QCarSim}.

\bmhead{Acknowledgements}
The authors acknowledge the AIMBeyonD Lab, in collaboration with which this research was conducted, and Heriot-Watt University Dubai for providing laboratory facilities and support.

\bmhead{Funding}
No funding was received for this study.

\bmhead{Competing interests}
The authors declare no competing interests.

\bmhead{Ethics approval and consent to participate}
Not applicable. This study did not involve human participants or animals.

\bmhead{Consent for publication}
Not applicable.

\bmhead{Author contributions}
M. Saeed: Conceptualization, Methodology, Software, Investigation, Formal analysis, Data curation, Visualization, Writing -- original draft, Writing -- review \& editing. M. Saleh: Supervision, Writing -- review \& editing. M. Shawgi: Supervision, Project administration, Writing -- review \& editing. M. Al Musleh: Resources, Writing -- review \& editing. All authors read and approved the final manuscript.

\bmhead{Thesis reuse disclosure}
This article builds upon the first author's undergraduate final-year project, ``Design and Evaluation of a Scalable Autonomous Vehicle Navigation Stack'', submitted to Heriot-Watt University Dubai in 2026.
\begin{appendices}

\section{Supplementary KPIs}\label{app:all_kpis}
\renewcommand{\theHtable}{\thesection.\arabic{table}}

\begin{table*}[!htbp]
\centering
\caption{Supplementary performance metrics for each
planner--controller combination (mean $\pm$ SD)}
\label{tab:all_kpis}

\footnotesize
\setlength{\tabcolsep}{4pt}

\begin{tabular}{@{}llcccc@{}}
\toprule
Controller & Planner
& \begin{tabular}[b]{@{}c@{}}
Path\\efficiency
\end{tabular}
& \begin{tabular}[b]{@{}c@{}}
Average speed\\(\si{\metre\per\second})
\end{tabular}
& \begin{tabular}[b]{@{}c@{}}
Final goal error\\(\si{\metre})
\end{tabular}
& \begin{tabular}[b]{@{}c@{}}
Heading error\\RMS (\si{\degree})
\end{tabular} \\
\midrule

\multicolumn{6}{@{}l}{\textit{Simulation}} \\

RPP & Hybrid A*
& $0.9621 \pm 0.000513$
& $0.186 \pm 0.00183$
& $0.1413 \pm 0.00173$
& $7.81 \pm 0.0124$ \\

& NavFn
& $0.9514 \pm 0.000374$
& $0.193 \pm 0.000259$
& $0.1400 \pm 0.00191$
& $18.3 \pm 0.0390$ \\

Stanley & Hybrid A*
& $0.9642 \pm 0.000555$
& $0.183 \pm 0.00208$
& $0.1377 \pm 0.00515$
& $6.87 \pm 0.0802$ \\

& NavFn
& $0.9650 \pm 0.00109$
& $0.161 \pm 0.000358$
& $0.1365 \pm 0.00953$
& $16.8 \pm 0.0616$ \\

Vector Pursuit & Hybrid A*
& $0.9160 \pm 0.00261$
& $0.222 \pm 0.00433$
& $0.1404 \pm 0.00431$
& $7.75 \pm 0.0386$ \\

& NavFn
& $0.9290 \pm 0.00115$
& $0.213 \pm 0.00130$
& $0.1407 \pm 0.00142$
& $20.7 \pm 0.0932$ \\

\midrule
\multicolumn{6}{@{}l}{\textit{Hardware}} \\

RPP & Hybrid A*
& $0.9984 \pm 0.00291$
& $0.674 \pm 0.0237$
& $0.07667 \pm 0.0644$
& $7.36 \pm 0.356$ \\

& NavFn
& $0.9978 \pm 0.0126$
& $0.509 \pm 0.0372$
& $0.05500 \pm 0.0276$
& $10.1 \pm 1.35$ \\

Stanley & Hybrid A*
& $1.006 \pm 0.00649$
& $0.445 \pm 0.0250$
& $0.07285 \pm 0.0100$
& $9.32 \pm 0.957$ \\

& NavFn
& $0.9901 \pm 0.00902$
& $0.398 \pm 0.00883$
& $0.08353 \pm 0.00996$
& $7.48 \pm 1.21$ \\

Vector Pursuit & Hybrid A*
& $1.032 \pm 0.00993$
& $0.631 \pm 0.0340$
& $0.09026 \pm 0.0560$
& $10.2 \pm 0.600$ \\

& NavFn
& $1.045 \pm 0.0364$
& $0.493 \pm 0.0625$
& $0.08126 \pm 0.0282$
& $15.0 \pm 4.13$ \\

\bottomrule
\end{tabular}

\par\smallskip
\begin{minipage}{\textwidth}
\footnotesize
\emph{Note:} Simulation average speeds
were calculated using simulation-clock time. They are below the
\SI{0.30}{\metre\per\second} simulation setpoint because the curvature-
and steering-based speed regulation (Section~\ref{sec:methods}) reduced the
commanded speed along the curved sections of the route.
\end{minipage}

\end{table*}
\end{appendices}

\bibliography{Bibliography}

\end{document}